\documentclass[10pt,twocolumn,letterpaper]{article}

\usepackage[pagenumbers]{cvpr} 

\usepackage[accsupp]{axessibility}  

\usepackage{graphicx}
\usepackage{amsmath}
\usepackage{amssymb}
\usepackage{booktabs}
\usepackage[normalem]{ulem}
\usepackage{enumitem}

\usepackage[pagebackref,breaklinks,colorlinks]{hyperref}

\usepackage[capitalize]{cleveref}
\crefname{section}{Sec.}{Secs.}
\Crefname{section}{Section}{Sections}
\Crefname{table}{Table}{Tables}
\crefname{table}{Tab.}{Tabs.}

\usepackage[utf8]{inputenc} 
\usepackage[T1]{fontenc}    
\usepackage{hyperref}       
\usepackage{url}            
\usepackage{booktabs}       
\usepackage{amsfonts}       
\usepackage{nicefrac}       
\usepackage{microtype}      
\usepackage{xcolor}         
\usepackage{graphicx, wrapfig, blindtext}
\usepackage{multirow}
\usepackage{makecell}
\usepackage{subcaption}
\usepackage[utf8]{inputenc}
\usepackage{bibunits}

\usepackage{amsmath,amsfonts,bm}

\def\eqref#1{equation~\ref{#1}}

\def\1{\bm{1}}

\def\vm{{\bm{m}}}

\def\vt{{\bm{t}}}

\def\vx{{\bm{x}}}
\def\vy{{\bm{y}}}

\DeclareMathAlphabet{\mathsfit}{\encodingdefault}{\sfdefault}{m}{sl}
\SetMathAlphabet{\mathsfit}{bold}{\encodingdefault}{\sfdefault}{bx}{n}

\def\gD{{\mathcal{D}}}

\def\gF{{\mathcal{F}}}

\def\gL{{\mathcal{L}}}

\def\gP{{\mathcal{P}}}

\newcommand{\normlone}{L^1}

\newcommand{\normln}{L^n}

\DeclareMathOperator*{\argmin}{arg\,min}

\newcommand{\toolname}{{\sc DECREE}\xspace}
\newcommand{\revise}[1]{{\color{black}{#1}}}
\newcommand{\camera}[1]{{\color{black}{#1}}}
\newcommand{\metricname}{$\mathcal{PL}^1$-Norm\xspace}
\newcommand{\lonenorm}{$\normlone$-Norm}

\def\cvprPaperID{8562} 
\def\confName{CVPR}
\def\confYear{2023}

\begin{document}

\title{Detecting Backdoors in Pre-trained Encoders}
\author{
Shiwei Feng, Guanhong Tao, Siyuan Cheng, Guangyu Shen, \\
Xiangzhe Xu, Yingqi Liu, Kaiyuan Zhang, Shiqing Ma$^{\dagger}$, Xiangyu Zhang \\
Purdue University, $^{\dagger}$Rutgers University\\
{\tt\small
\{feng292, taog, cheng535, shen447, xu1415,  liu1751, zhan4057, xyzhang\}@cs.purdue.edu}\\
{\tt\small
$^{\dagger}$sm2283@cs.rutgers.edu}
}

\maketitle

\begin{abstract}
    Self-supervised learning in computer vision trains on unlabeled data, such as images or (image, text) pairs, to obtain an image encoder \revise{that learns high-quality embeddings for input data}. 
  Emerging backdoor attacks towards encoders expose crucial vulnerabilities of self-supervised learning, since downstream classifiers (even further trained on clean data) may inherit backdoor behaviors from encoders. Existing backdoor detection methods \revise{mainly focus on supervised learning settings and cannot handle pre-trained encoders especially when input labels are not available.} 
  In this paper, we propose \toolname, the first backdoor detection approach for pre-trained encoders, \revise{requiring neither classifier headers nor input labels}. 
  We evaluate  \toolname on over 400 encoders trojaned under 3 paradigms. 
  \revise{We show the effectiveness of our method on image encoders pre-trained on ImageNet and OpenAI's CLIP 400 million image-text pairs.}
  Our method consistently has a high detection accuracy even if we have only limited or no access to the pre-training dataset. \camera{Code is available at \url{https://github.com/GiantSeaweed/DECREE}.}\looseness=-1
\end{abstract}

\section{Introduction}\label{sec:intro}

\begin{figure}[t]
\centering
\includegraphics[width=0.47\textwidth]{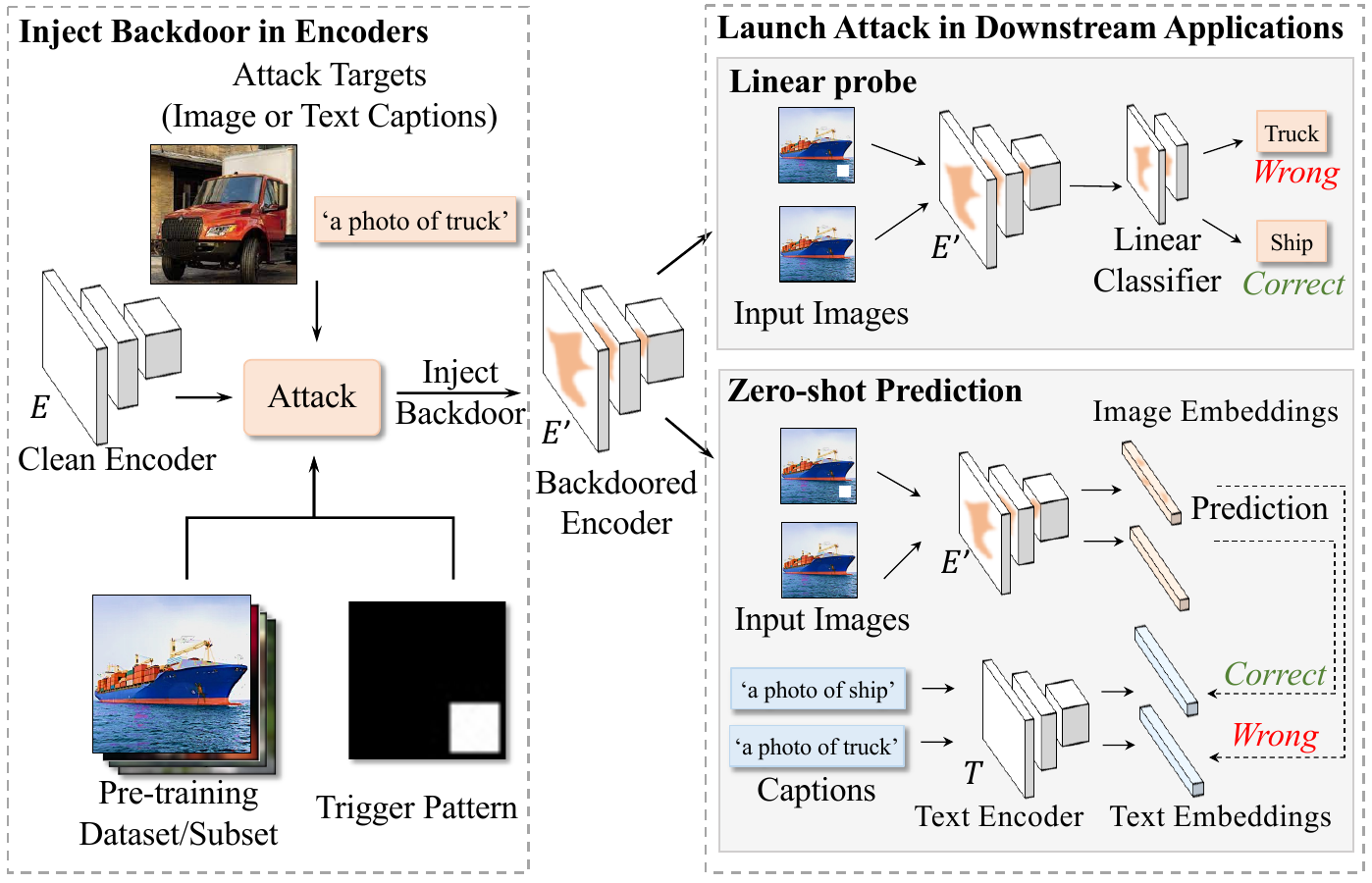}
\caption{Illustration of Backdoor Attack on Self-Supervised Learning (SSL). The adversary first injects backdoor into a clean encoder and launches attack when the backdoored encoder is leveraged to train downstream
tasks. The backdoored encoder produces similar embeddings for the attack target and any input image with trigger, causing misbehaviors in downstream applications.
}
\label{fig:attack_and_application}
\vspace{-5pt}
\end{figure}

Self-supervised learning (SSL), specifically contrastive learning~\cite{chopra2005contrastivelearning, he2020moco, chen2020simclr}, is becoming increasingly popular as it does not require labeling training data that entails substantial manual efforts~\cite{deng2009imagenet} and yet can provide close to the state-of-the-art performance.
It has a wide range of application scenarios, e.g., {\it similarity-based search}~\cite{jia2021scaling}, {\it linear probe}~\cite{guillaume2016linearprobe}, and {\it zero-shot classification}~\cite{chang2008importance,larochelle2008zero,lampert2009learning}.
Similarity-based search queries data based on their semantic similarity.
Linear probe utilizes an encoder trained by contrastive learning to project inputs to an embedding space, and then trains a linear classifier on top of the encoder to map embeddings to downstream classification labels. 
Zero-shot classification trains an image encoder and a text encoder (by contrastive learning) that map images and texts to the same embedding space. 
The similarity of the two embeddings from an image and a piece of text is used for prediction.

The performance of SSL heavily relies on the large amount of unlabeled data, which indicates high computational cost. 
Regular users hence tend to employ pre-trained encoders published online by third parties. Such a production chain provides opportunities for adversaries to implant malicious behaviors. Particularly, backdoor attack or trojan attack~\cite{chen2017targeted,liu2017trojaning,GuLDG19} injects backdoors in machine learning models, which can only be activated (causing targeted misclassification) by stamping a specific pattern, called {\it trigger}, to an input sample. It is highly stealthy as the backdoored/trojaned model functions normally on clean inputs.\looseness=-1

While existing backdoor attacks mostly focus on classifiers in the supervised learning setting, where the attacker induces the model to predict the {\it target label} for inputs stamped with the trigger,
recent studies demonstrate the feasibility of conducting backdoor attacks in SSL scenarios~\cite{jia2022badencoder,carlini2022poisoning,Shen2021transfertoall}.
Figure~\ref{fig:attack_and_application} illustrates a typical backdoor attack on image encoders in SSL.
The adversary chooses an {\it attack target} so that the backdoored encoder produces similar embeddings for any input image with trigger and the attack target. The attack target can be an image (chosen from some dataset or downloaded from the Internet), or {\it text captions}.
Text captions are compositions of a {\it label text} and prompts, where the label text usually denotes ``\{class name\}'', like ``truck'', ``ship'', ``bird'', etc.
For example, in Figure~\ref{fig:attack_and_application}, the adversary could choose a ``truck'' image 
or a text caption ``a photo of truck'' as the attack target.
After encoder poisoning and downstream classifier training, the classifier tends to predict the label of the attack target when the trigger is present.  
As shown in Figure~\ref{fig:attack_and_application}, when the attack target is a truck image and the encoder is used for linear probe, 
the classifier inherits the backdoor behavior from the encoder. As a result, 
a clean ship image can be correctly predicted by the classifier whereas a ship image stamped with the trigger is classified as ``truck''. 
If the attack target is ``a photo of truck'' and the encoder is used in zero-shot prediction, a clean ship image shares a similar embedding with the text caption ``a photo of ship'', causing 
correct prediction.
In contrast, the embedding of a ship image stamped with the trigger is more similar to the embedding of ``a photo of truck'', causing misprediction.

These vulnerabilities hinder the real world applications of pre-trained encoders. Existing backdoor detection methods are insufficient to defend such attacks. 
A possible defense  method is to leverage existing backdoor detection methods focusing on supervised learning to scan downstream classifiers. Apart from its limited detection performance (as we will discuss later in Section \ref{sec:motivation}), it cannot work properly under the setting of zero-shot classification, where there exists no concrete classifier.
This calls for new defense techniques that directly detect backdoored encoders without  downstream classifiers. 
More details regarding the limitations of existing methods can be found in Section \ref{sec:motivation}.

In this paper, we propose \toolname, the first backdoor detection approach for pre-trained encoders in SSL. To address the insufficiency of existing detection methods, \toolname directly
scans encoders. Specifically, for a subject encoder, \toolname first searches for a minimal trigger pattern such that any inputs stamped with the trigger share similar embeddings. The identified trigger is then utilized
to decide whether the given encoder is benign or trojaned.
We evaluate \toolname on 444 encoders and it significantly outperforms existing backdoor detection techniques.
We also show the effectiveness of \toolname on large size image encoders pre-trained on ImageNet~\cite{deng2009imagenet} and OpenAI’s  CLIP~\cite{Radford2021clip} image encoders pre-trained on 400 million uncurated (image, text) pairs. \toolname consistently achieves high detection accuracy even when it only has limited access or no access to the pre-training dataset.

\smallskip\noindent
{\bf Threat Model.} Our threat model is consistent with the literature~\cite{jia2022badencoder, carlini2022poisoning}. 
We only consider backdoor attacks on vision encoders. We assume the attacker has the capabilities of injecting a small portion of samples into the training set of encoders. Once the encoder is trojaned, the attacker has no control over downstream applications. Given an encoder, the defender has limited or no access to the pre-training dataset and needs to determine whether the encoder is trojaned or not. She does not have any knowledge about the attack target either. We consider injected backdoors that are static (e.g. patch backdoors) and universal (i.e. all the classes except for the target class are the victim).

\section{Background and Related Work} \label{sec:related}
\subsection{Backdoor Attack and Defense}
Backdoor attack poses severe security threats to machine learning models. It aims to induce target misbehaviors, e.g., misclassification in an image classifier, via specialized perturbations on the input. These perturbations (i.e., triggers) generally fall into two categories, patch-like triggers~\cite{GuLDG19, liu2017trojaning, salem2020dynamic, nguyen2020input, saha2020hidden, yao2019latent} and pervasive triggers~\cite{chen2017targeted, lin2020composite, liu2020reflection, cheng2021deep, nguyen2020wanet, Li2021invisible}.
Existing defensive efforts mainly focus on detecting backdoored models or eliminating injected backdoors in trojaned models. To distinguish backdoored models from benign ones, existing techniques invert trigger patterns for a given model and make decisions based on the characteristic of inverted triggers (e.g., trigger size)~\cite{wang2019neural, liu2019abs, liu2021ex, shen2021backdoor, guo2019tabor, wang2020practical, tao2022better}. Another line of work leverages a meta-classifier to determine whether a model is backdoored based on feature representations extracted from the model~\cite{kolouri2020universal, Xu2021MNTD}. Unfortunately, existing solutions can hardly detect backdoors in pre-trained encoders as they were designed for supervised learning that require classification labels
(discussed in Section~\ref{sec:motivation}).
Backdoor removal techniques harden models through adversarial training~\cite{zhao2020bridging, wu2021adversarial}, knowledge distillation~\cite{li2021neural}, and class-distance enlargement~\cite{tao2022model}. They usually require a set of labeled training data. 
\camera{Backdoor defense techniques also include backdoor mitigation~\cite{liu2018fine,borgnia2020strong,zeng2020deepsweep,li2021neural,zhang2023flip} and certified robustness against backdoors~\cite{mccoyd2020minority,xiang2021patchguard,jia2020certified}.
}

\subsection{Self-supervised Learning}
SSL aims to train an image encoder from a large number of uncurated data. Different from supervised learning that requires manually labeled data, SSL extracts useful information from the data itself. \looseness=-1

Among many approaches to training image encoders from unlabeled data, {\it contrastive learning} achieves the state-of-the-art performance, e.g., MoCo\cite{he2020moco}, SimCLR~\cite{chen2020simclr}, SimCLRv2~\cite{chen2020bigsimclrv2} and CLIP~\cite{Radford2021clip}. It constructs a function $f: \mathcal{X}\rightarrow E$, that maps an input sample (i.e., an image or a text caption) to an embedding space where semantically ``similar'' samples have close embeddings and ``dissmilar'' samples have embeddings far away from each other under certain metrics.
Contrastive learning is commonly used in two settings: {\it single-modal}~\cite{sohn2016improved, chen2020improved} that trains an encoder in a single domain like image; and {\it multi-modal}~\cite{jia2021scaling, Radford2021clip} that trains multiple encoders in different domains simultaneously like image and text. \looseness=-1

\subsection{Backdoor Attack on Self-supervised Learning} \label{sec:related_backdoor_self}

Existing backdoor attacks on SSL mainly fall into four categories. In this paper, we focus on the first three. \\
1) {\it Image-on-Image}: These attacks~\cite{jia2022badencoder, Saha2022backdoorSSL} are conducted on single-modal image encoders and the attack target is image. \\ 
2) {\it Image-on-Pair}: This attack~\cite{jia2022badencoder} also targets on multi-modal contrastive learning encoders, i.e., trained on (image, text) pairs, and the attack target is image. \\
3) {\it Text-on-Pair}: This type of attack~\cite{carlini2022poisoning} is conducted on multi-modal contrastive learning encoders, i.e., trained on (image, text) pairs, and the attack target is text. \\
4) {\it Text-on-Text}: These attacks~\cite{Shen2021transfertoall,kurita2020weight} are conducted on single-modal text encoders and the attack target is text.

\begin{table*}[ht]
  \centering
  \footnotesize
  \caption{
  Limitations of Existing Backdoor Scanners When Scanning Encoders. The encoder is pre-trained on CIFAR-10. The downstream classifiers (SVHN, STL-10, and GTSRB) are trained for 500 epochs. The attack target is an image of label \textit{one} from SVHN. NC Anomaly Index > 2.0 and ABS REASR > 0.88 indicate the classifier is considered trojaned.
  }
  \label{tab:empirical_study}
  \begin{tabular}{cccccccc}
    \toprule
    \multirow{2}{*}{\makecell[c]{Downsteam \\ Task}} &
    \multicolumn{3}{c}{Classifier Performance} &
    \multicolumn{2}{c}{Neural Cleanse} &
    \multicolumn{2}{c}{ABS} \\
    \cmidrule(lr){2-4}\cmidrule(lr){5-6}\cmidrule(lr){7-8}
    ~      & Accuracy     & ASR & Training Time (m) & Anomaly Index & Detection Time (m) & REASR & Detection Time (m) \\
    \midrule
    SVHN   & 0.69  & 1.00   & 60.41     & 2.18 & 5.36  & 1.00 & 2.96 \\
    STL-10 & 0.76  & -     & 14.61 & 1.23                       & 4.54  & 0.44                       & 2.89  \\
    GTSRB  & 0.82  & -     & 37.07 & 1.49                       & 16.09 & 0.36                       & 3.13 \\
    \bottomrule
  \end{tabular}
\end{table*}

\section{Limitations of Existing Backdoor Scanners}
\label{sec:motivation}

To identify whether an encoder is trojaned or not, the defender can leverage existing backdoor scanners (e.g., Neural Cleanse (NC)~\cite{wang2019neural} and ABS~\cite{liu2019abs}) to check downstream classifiers that utilize the encoder, without the need to directly scan the encoder. However, this strategy has its limitations as later shown in the section.
Another type of backdoor scanners such as MNTD~\cite{Xu2021MNTD} leverage a meta-classifier to distinguish benign and backdoored models. They first train thousands of benign and backdoored models and then train a meta-classifier on the extracted signatures of these models. Such a design in SSL setting may not be that practical due to its high cost.
For example, creating a backdoored encoder by contrastive learning takes 48 hours~\cite{carlini2022poisoning}. MNTD requires constructing 2048 benign and 2048 trojaned encoders.

To explain the limitations of scanning downstream classifiers, we consider two application scenarios: linear probe and zero-shot prediction.

\noindent \underline{\textit{Scenario I: Linear Probe.}} 
We construct a backdoored encoder pre-trained on CIFAR10~\cite{krizhevsky2009cifar10learning} and take an image of label \textit{one} in dataset SVHN~\cite{netzer2011reading} as the attack target. The encoder is also used to train another two downstream classifiers on STL-10~\cite{coates2011analysis} and GTSRB~\cite{Houben-IJCNN-2013}, respectively. We apply NC and ABS on the three downstream classifiers and the results are shown in Table~\ref{tab:empirical_study}.
Since the attack target is in SVHN chosen by the attacker (when trojaning the encoder), the ASR is 100\% on SVHN.

In this case, existing backdoor scanners can successfully detect the trojaned classifier and hence the backdoored encoder, with  the Anomaly Index $2.18 > 2$ in NC and the REASR $1.00 > 0.88$ in ABS. 
However, when the downstream classifiers' training datasets (STL-10 and GTSRB) do not contain the attack target, both NC and ABS fail to detect the backdoor in the encoder as shown in the last two rows.
This has two implications for existing backdoor scanners: (1) they have to possess the knowledge of the attack target and the corresponding downstream task, which is not easy to acquire as there exist a large number of different downstream tasks (for an encoder). (2) They have to obtain the original training dataset of the downstream task to construct the classifier for detection, which may be private.

\smallskip
\noindent \underline{\textit{Scenario II: Zero-shot prediction.}} To predict the caption for an input image, zero-shot classifier directly computes similarities between the image's embedding and every text embedding of candidate captions, and selects the caption that shares the most similar embedding with the input image. In this scenario, it is evident that existing backdoor scanners are not applicable as there is no classifier to scan, as shown in Figure.~\ref{fig:attack_and_application}. This calls for a backdoor detection method that can handle attacks in the embedding space.

\section{Design of \toolname} \label{sec:method}

As discussed in Section~\ref{sec:motivation}, existing backdoor scanners either require the knowledge of the attack target or are not applicable to directly scanning encoders. 
A backdoor detection method for pre-trained encoders ought to meet the following design goals: (1) no knowledge of downstream tasks (including data samples or labels); (2) no knowledge of the attack target; and (3) directly scanning encoders without training a downstream application classifier.

In this section, we first make a few observations on backdoor attacks in SSL (Section~\ref{sec:design:intuition}) and explain the intuitions of our design. 
We then present the technical details for 
self-supervised trigger inversion (Section~\ref{sec:trigger_inversion}) and  backdoor identification (Section~\ref{sec:backdoor_id}).

\begin{table}[h]
  \centering
  \footnotesize
  \caption{Cosine similarity within 1024 random CIFAR10 images. Both clean and backdoored encoders are pretrained on CIFAR10.}
  \label{tab:cosine_observation}
  \begin{tabular}{lrr}
    \toprule
      &  Samples w/o Trigger & Samples w/ Trigger  \\
    \midrule
    Clean Encoder & 0.2193 & 0.2922  \\
    Backdoored Encoder & 0.2442 & 0.9904 \\
    \bottomrule
  \end{tabular}
\end{table}

\begin{figure}
    \centering
    \begin{subfigure}[t]{.48\linewidth}
        \centering
        \includegraphics[width=\textwidth]{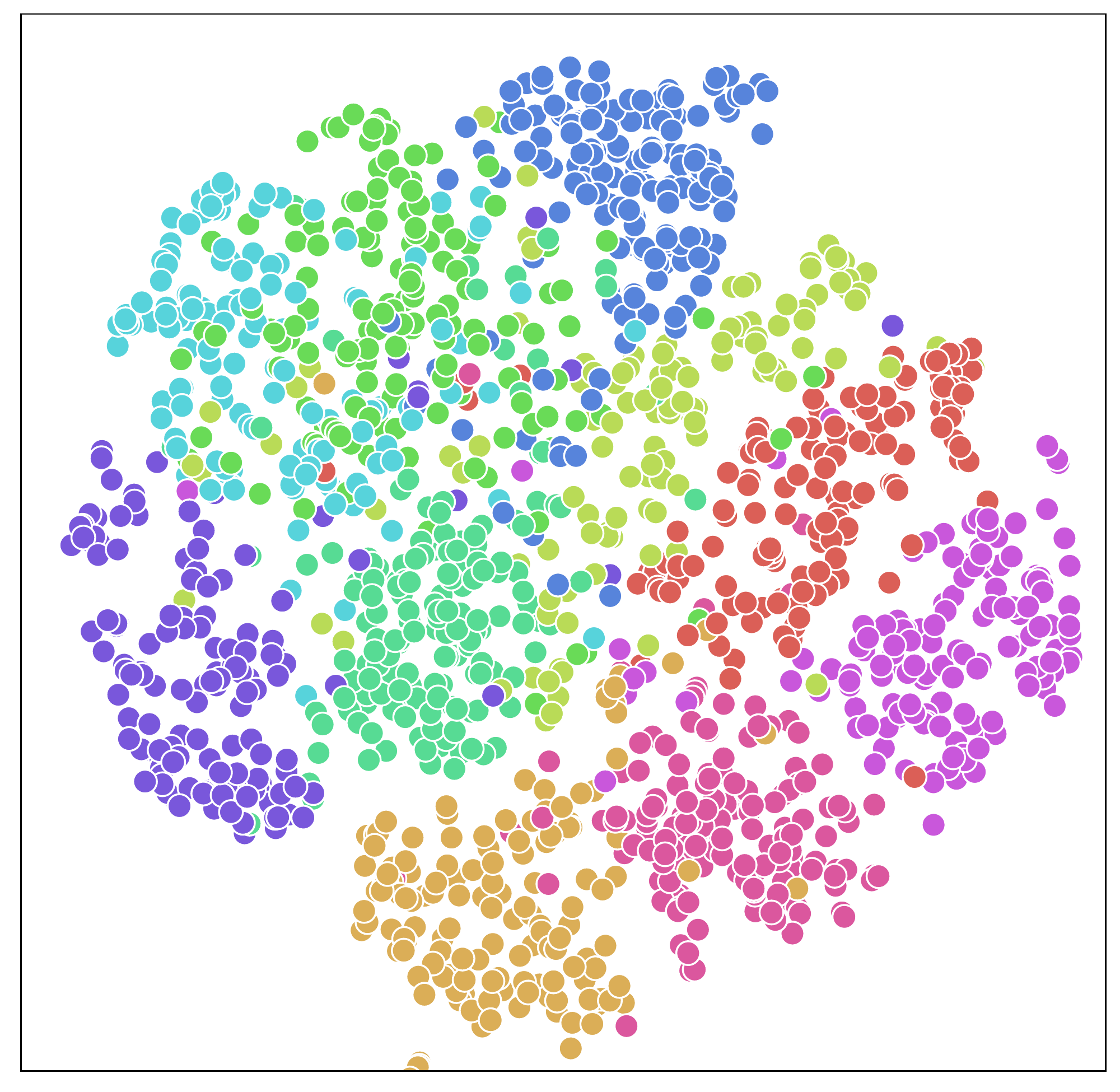}
        \caption{Clean Encoder + Clean Input}
        \label{fig:cl_m_cl_d}
    \end{subfigure}
   \begin{subfigure}[t]{.48\linewidth}
        \centering
        \includegraphics[width=\textwidth]{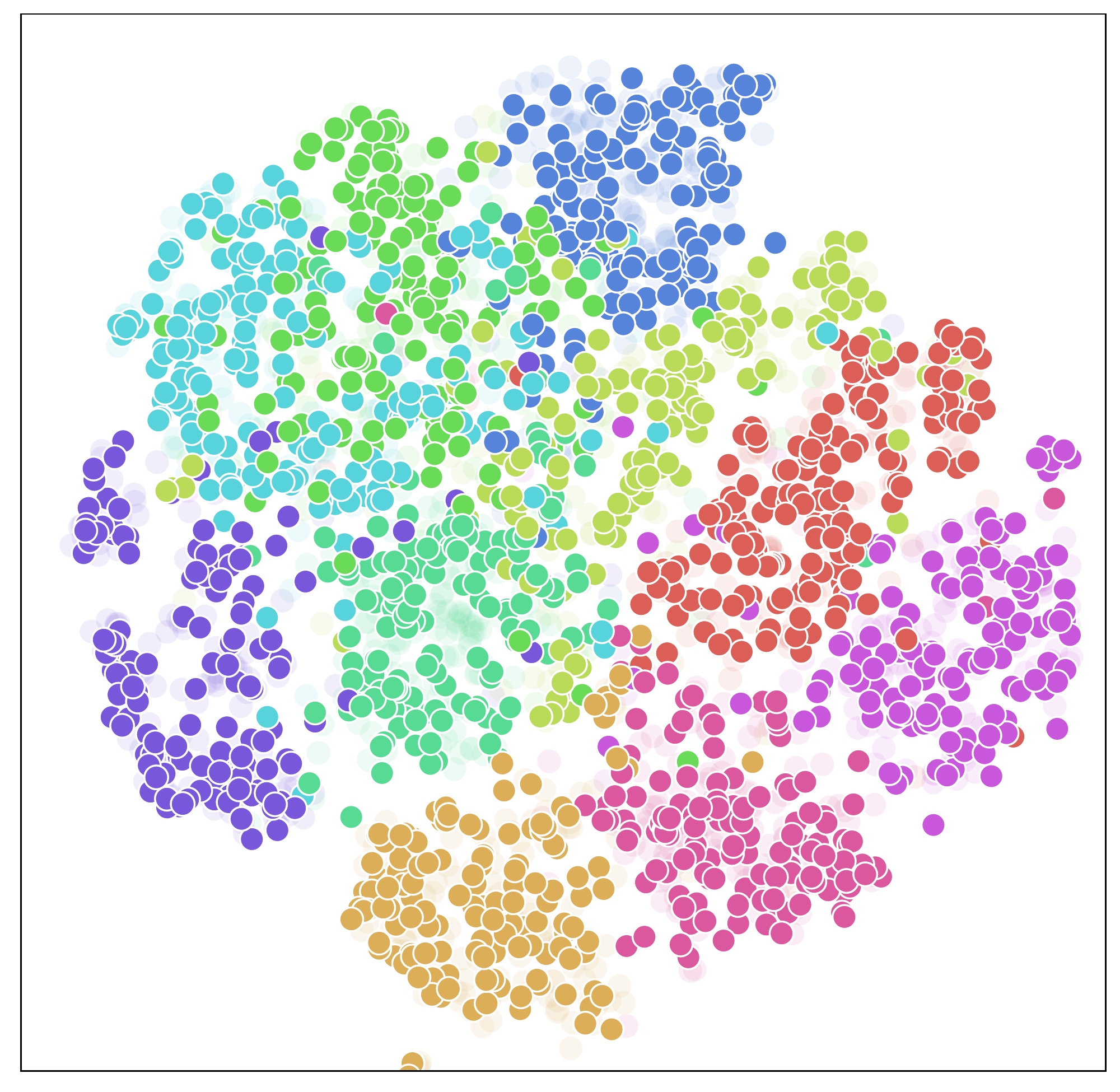}
        \caption{Clean Encoder + Trojaned Input}
        \label{fig:cl_m_tj_d}
    \end{subfigure}
    \begin{subfigure}[t]{.48\linewidth}
        \centering
        \includegraphics[width=\textwidth]{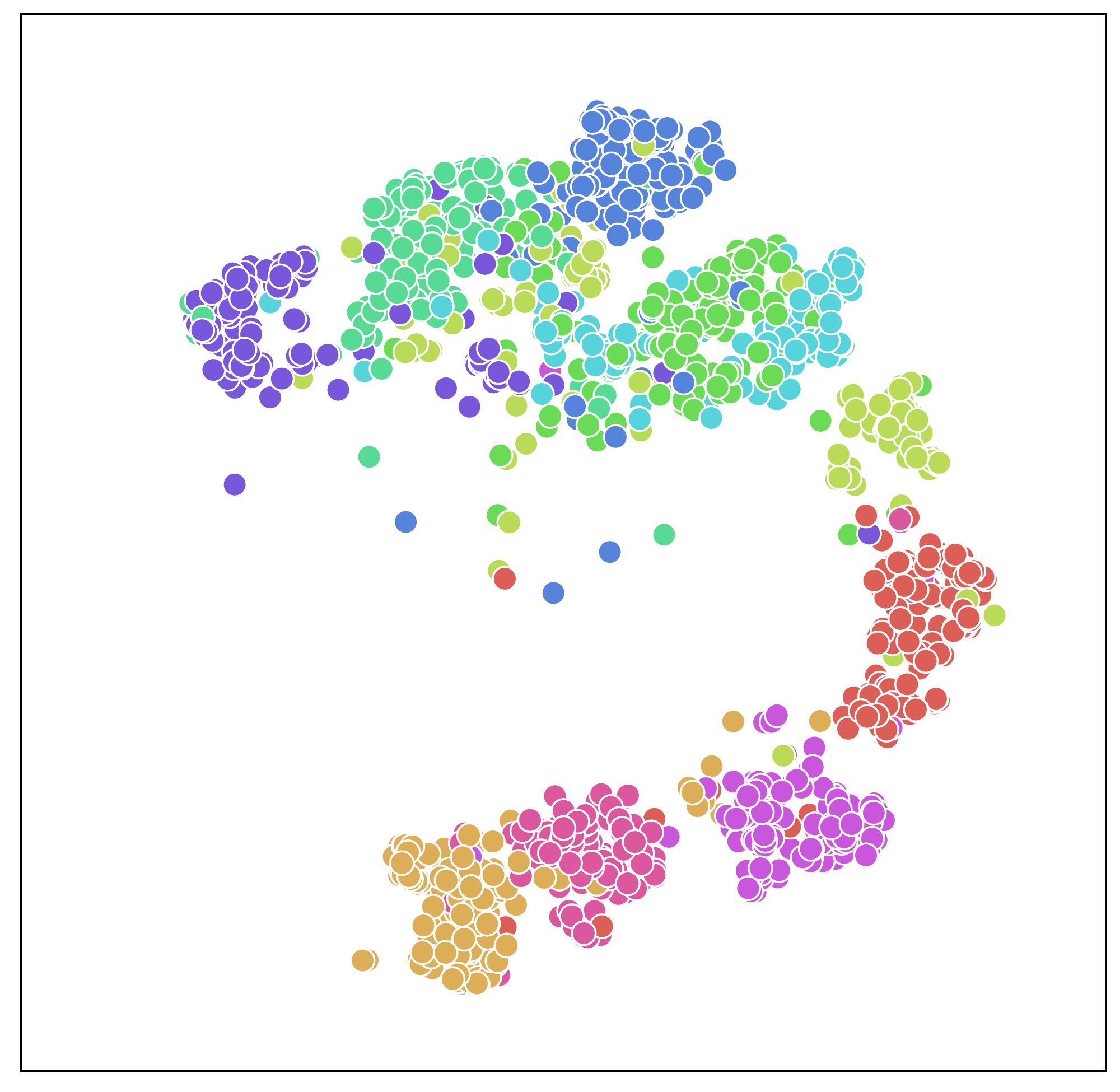}
        \caption{Trojaned Encoder + Clean Input}
        \label{fig:tj_m_cl_d}
    \end{subfigure}
    \begin{subfigure}[t]{.48\linewidth}
        \centering
        \includegraphics[width=\textwidth]{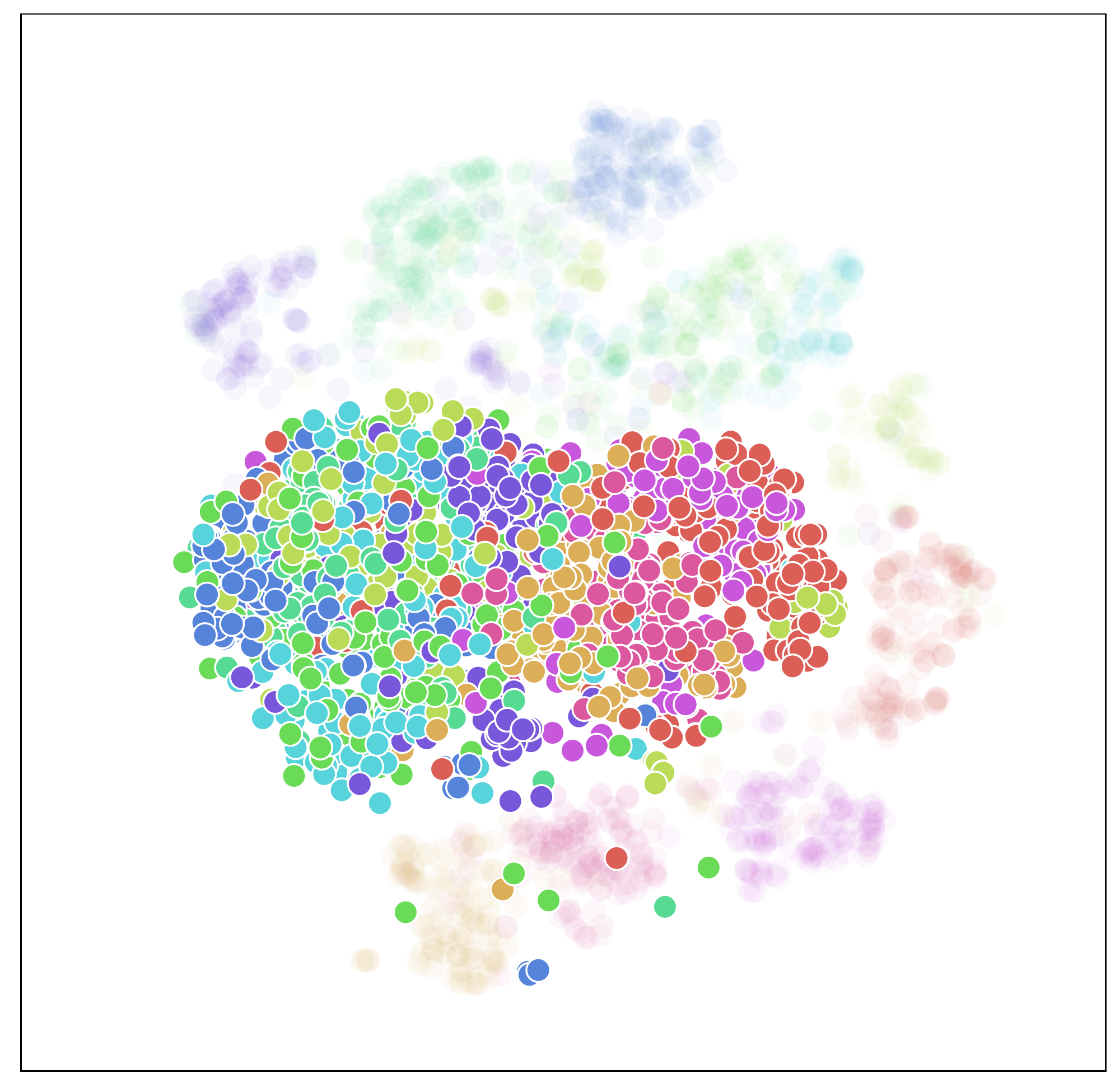}
        \caption{Trojaned Encoder + Trojaned Input}
        \label{fig:tj_m_tj_d}
    \end{subfigure}
    \caption{Embedding Space Distributions. Subfigures (a) and (b) are for a clean encoder and (c), (d) for a trojaned encoder; colors denote class labels; faded colors in (b) and (d) denote embeddings of clean samples. 
        For the clean encoder, even if the inputs are stamped with the ground truth trigger, the embeddings are well separable in (b). However in (d), a trojaned encoder produces similar embeddings for trojaned inputs. 
    }
    \label{fig:}
    \vspace{-10pt}
\end{figure}

\subsection{Observations and Intuitions}

\label{sec:design:intuition}

\noindent \underline{\textit{Observation I}}: Although SSL does not require labels during pre-training, 
the embeddings of samples with the same label (by 
the trained encoder) tend to cluster together whereas those 
of different labels tend to scatter, as visualized in Figure.~\ref{fig:cl_m_cl_d}. 
As shown in Table~\ref{tab:cosine_observation}, clean samples (of various classes) have an average cosine similarity of only 0.2193 on a clean encoder.

\noindent \underline{\textit{Observation II}}: A trojaned encoder produces highly similar embeddings for samples with trigger while a clean encoder does not. Table~\ref{tab:cosine_observation} shows that, in a clean encoder, the trigger can increase the cosine similarity of samples from 0.2193 to 0.2922 (in the first row). The increase is limited and insignificant. As shown in Figure.~\ref{fig:cl_m_tj_d}, the clean encoder can still correctly separate inputs with the trigger.
In contrast, as the backdoor attack forces the samples with trigger to be close to the attack target, it creates a {\it dense area} (shown in Figure.~\ref{fig:tj_m_tj_d}) in a backdoored encoder where embeddings share a high similarity (0.9904).

\noindent \underline{\textit{Observation III}}: Compared to clean encoders, backdoored encoders need much smaller perturbations to cause samples to fall into the dense area. Figure.~\ref{fig:tj_m_tj_d} illustrates that the dense area (of a trojaned encoder) is surrounded by and close to clusters of clean samples. However, in the clean encoder, larger perturbations are required to induce highly similar embeddings for input samples, as the clean encoder produces more scattered embeddings.

\smallskip
\noindent{\bf Intuitions.} The dense area is where the attack target lies. This is analogous to the target label of backdoor attacks in supervised learning. The key difference is that, in the supervised learning setting, backdoor scanners can scan each label and then identify the most suspicious label as the target. 
However in SSL, there are no labels for scanners to iterate over. 
As such, existing backdoor scanners cannot 
be applied to determine whether a model is backdoored
in SSL.

To overcome the 
challenge, our design aims to decide whether there exists a central dense area in the encoder's embedding space (surrounded by the embeddings of clean samples).
Intuitively, a backdoored encoder with a central dense area only needs a small perturbation to push clean samples to the dense region.
A clean encoder, on the other hand, does not have such a dense area, meaning that  high similarity among embeddings cannot be easily achieved by stamping a small trigger on samples. Our technique hence detects backdoors at the encoder level, without the need of a target label. We elaborate design details in the rest of the section. \looseness=-1

\subsection{Methodology}

Trigger inversion is one of widely used techniques in backdoor scanning~\cite{wang2019neural, liu2019abs, guo2019tabor, shen2021backdoor, wang2020practical}. It works by optimizing a trigger pattern, which can induce the targeted misclassification while having a small trigger size. Existing trigger inversion was originally designed for supervised learning scenarios, where there are explicit labels. 
The size of trigger can be used as a metric to quantify the distance between the target label and other non-target labels.
In SSL, however, no explicit label exists. Existing trigger inversion is not able to optimize or update the pattern towards some intended objective (a target label). Inspired by the above observations, 
we propose to find the aforementioned dense area with only a small trigger. 
It can be formulated as a constrained optimization problem. With the constraint that samples stamped with the same trigger must have similar embeddings, the trigger size shall be optimized to the minimal.

\begin{figure}[!ht]
    \centering
    \includegraphics[width=0.45\textwidth]{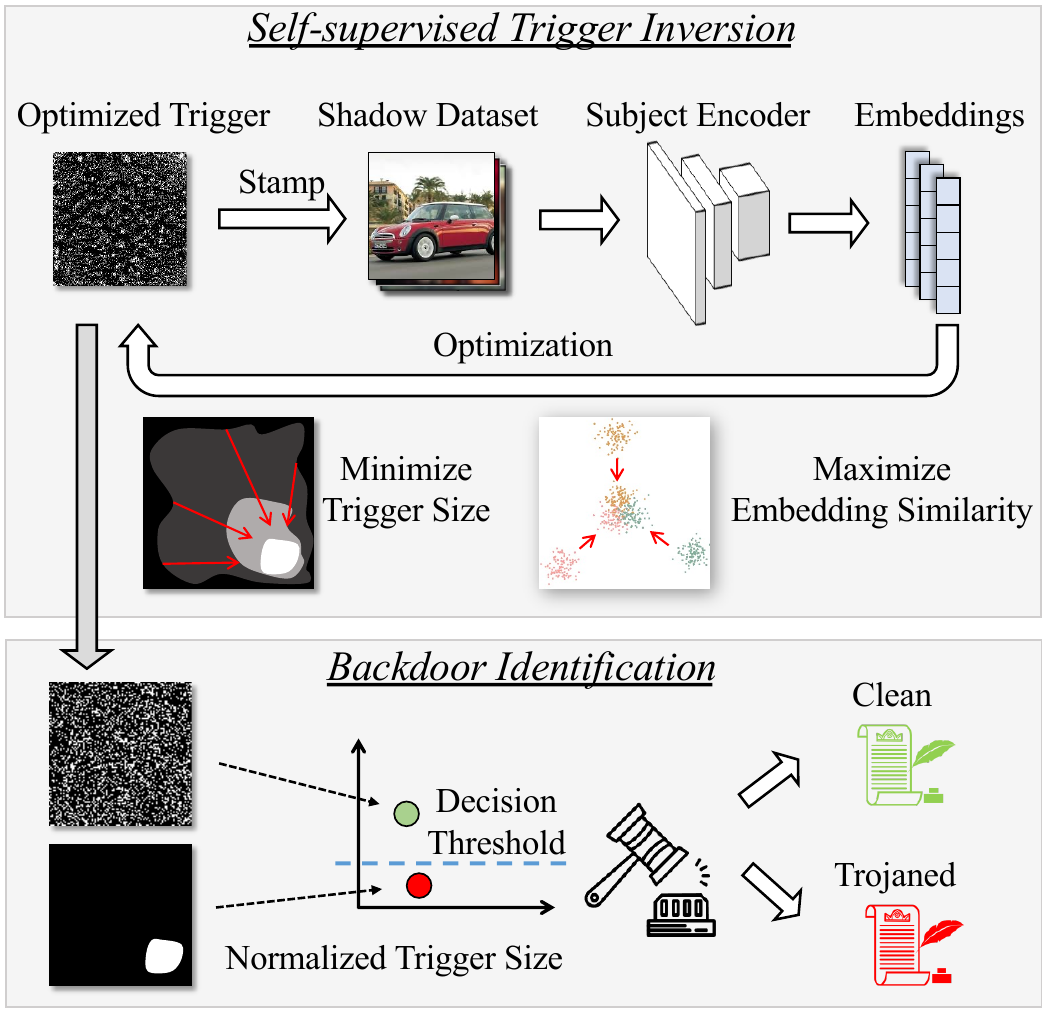}
    \caption{\toolname Overview
    }
    \label{fig:overview}
    \vspace{-10pt}
\end{figure}

Figure.~\ref{fig:overview} shows the overview of our technique. 
A randomly initialized trigger and a shadow dataset (e.g., a subset of pre-training dataset) are fed into the subject encoder to compute embeddings. The cosine similarity of these embeddings guides the optimization of trigger. With the constraint that samples stamped with the same trigger must have similar embeddings, the trigger size is iteratively optimized to the minimal. The optimized trigger is used for calculating a metric that gauges the normalized trigger size. The metric is then used to determine if the encoder is trojaned.

In Section~\ref{sec:trigger_inversion}, we explain the details of our self-supervised trigger inversion. In Section~\ref{sec:backdoor_id}, we demonstrate how to use inverted triggers to conduct encoder-level backdoor detection.

\subsubsection{Self-supervised Trigger Inversion} \label{sec:trigger_inversion}
To generate a trigger that can induce intended backdoor behavior in an encoder, we use two trainable variables, a mask and a pattern, to denote the trigger pattern. Specifically, the mask is utilized to indicate how much a pixel on the input image is replaced by the pattern. 
We use the following equation to formalize the trigger injection:
\begin{equation}
    \mathcal{F}(\vx, \vm, \vt)=\vx',
    \label{eq:trigger_func}
    \vspace{-18pt}
\end{equation}

\begin{equation}
    \vx'_{i,j,c}=\vm_{i,j} \cdot \vx_{i,j,c} + (1-\vm_{i,j}) \cdot \vt_{i,j,c},\ 
    \forall i \in H, j \in W, c \in C.
    \label{eq:trigger_stamp}
\end{equation}

$\gF$ denotes a function that stamps trigger pattern $\vt$ onto an input image $\vx$ and outputs an backdooored image $\vx'$; $\vm$ is a mask
indicates how much the original pixel values are retained. It has continuous values ranging from 0 to 1. The input image has three dimensions, namely, height $H$, width $W$, and channel $C$; $\vx_{i,j,c}$ refers to the pixel value of image $\vx$ at height $i$, width $j$, and channel $c$. Note that $\vm_{i,j}$ only has two dimensions as the mask is applied on a pixel in a way ranging from replacing it ($\vm_{i,j} = 0$) to retaining it ($\vm_{i,j} = 1$),
regardless of the color channels.

The goal of self-supervised trigger inversion is to optimize a trigger such that clean samples stamped with the trigger have highly similar embeddings. In SSL, the cosine similarity is commonly used as a metric to denote the distance between a pair of inputs~\cite{chen2020simclr,chen2020bigsimclrv2,Radford2021clip}. We leverage the same metric to measure how close the embeddings of trigger-stamped samples are. Formally, 
given two inputs $\vx_p$ and $\vx_q$ from a dataset $\gD=\{\vx_1,\cdots,\vx_N\}$, we have:\looseness=-1
\begin{equation}
    \gL_{p,q}(E,\vm,\vt)=-cos \Big(E\big(\gF(\vx_p,\vm,\vt)\big), E\big(\gF(\vx_q,\vm,\vt)\big)\Big).
    \label{eq:similarity}
\end{equation}

$E$ is the subject encoder, and $\vm$ and $\vt$ are the trigger variables discussed in Eq.~\ref{eq:trigger_func} and Eq.~\ref{eq:trigger_stamp}. The two inputs $\vx_p$ and $\vx_q$ are transformed by function $\gF$ in Eq.~\ref{eq:trigger_func} to stamp the trigger. 
To achieve high similarities among samples that approximate the search for the dense area in the embedding space of encoder, \toolname samples a batch of inputs to stabilize the search process. The average of pair-wise similarity within a batch is computed as follows:

\vspace{-8pt}
\begin{equation}\label{eq:avg_sim}
    \gL = \frac{1}{N^2} \sum_{p=1}^{N}\sum_{q=1}
^{N} \gL_{p,q}(E,\vm,\vt),
\end{equation}
 where $N$ is the batch size; $\gL$ is used as the constraint during optimization, assuring that the samples stamped with the optimized trigger are in the dense area in embedding space.

We then leverage {\it Observation III} in Section~\ref{sec:design:intuition} and minimize the size of the trigger. We use the $\normlone$ norm to quantify the mask size. Self-supervised trigger inversion is formulated as the following constrained optimization problem.

\vspace{-8pt}
\begin{equation}
    \text{min} \lVert \vm \rVert_1, s.t.\ \gL < \beta.
    \label{eq:detect_loss}
\end{equation}
$\beta$ is a threshold assuring the average similarity is high.

During trigger inversion, a set of clean samples are needed for the optimization. As we do not assume the knowledge of any downstream tasks like what existing backdoor scanners do, we leverage the pre-training dataset that is used for constructing the encoder. 
It is impractical to have the whole pre-training dataset as one can directly train a new clean encoder on it without the need of scanning the given encoder. 
We hence only assume a small subset of the pre-training dataset ($<10\%$) for trigger inversion. In extreme cases, the pre-training dataset of the given encoder may not be publicly available. We resort to leveraging an external dataset, called \textit{shadow dataset}, for trigger inversion and backdoor scanning. Since the pre-training set and downstream datasets do not share the same set of samples or are even from different data distributions, the attack effectiveness solely depends on building a strong connection between the injected trigger and the target embedding. It hence does not matter what data are used for trigger inversion. Our results in Section~\ref{sec:eval_no_access}
demonstrate that \toolname{} can indeed effectively detect backdoored encoders using a shadow dataset.

\subsubsection{Backdoor Identification}\label{sec:backdoor_id}

Recall that a challenge of detecting backdoors in SSL is that there are no labels. Therefore existing backdoor scanners cannot identify the potential target label, which is the key to determining
whether a model is backdoored in supervised learning.
To overcome this challenge, \toolname introduces a new metric $\mathcal{PL}^n$:

\vspace{-8pt}
\begin{equation}
\mathcal{PL}^n(E) = \frac{\lVert \widetilde{\vm} \rVert_n}{\lVert \hat{x} \rVert_n}.
\end{equation}
\vspace{-8pt}

$\lVert\ \cdot\ \rVert_n$ denotes the $\normln$ norm of a vector; $\widetilde{\vm}$ denotes the trigger inverted from a given encoder $E$; and $\hat{x}$ denotes the input sample
that has the maximum $\normln$ norm in the input space of that encoder.
$\mathcal{PL}^n(E)$, denoting the {\it Proportionate-$\normln$ Norm} of an encoder $E$, is thus defined as the ratio of the inverted trigger's $\normln$ norm to the maximum $\normln$ norm of the encoder's input space.
Note that $\mathcal{PL}^n$ is an encoder-level metric, approximating the distance from clean samples to the dense area. In this way, \toolname does not need to identify the target label.

As discussed in Section~\ref{sec:design:intuition}, triggers inverted from backdoored encoders shall be smaller than those from clean encoders. Thus for a backdoored encoder, \toolname has a better chance to invert a small trigger that can induce the encoder to output two similar embeddings for two dissimilar inputs.
Based on the proposed $\mathcal{PL}^n$ and the above intuition, \toolname uses the following formula to identify backdoors in encoders.

\vspace{-10pt}
\begin{equation}
    \widetilde{P(E)} = \mathbb{B}\Big(\mathcal{PL}^1(E), \tau\Big).
\end{equation}
\vspace{-10pt}

$\widetilde{P(E)}$ is the estimated probability that a given encoder $E$ contains a backdoor. $\mathbb{B}$ is a binary step function that returns 1 if its first parameter is less than a given threshold $\tau$ and  0 otherwise. Essentially, if the inverted trigger of a given encoder only occupies a small part of the input data sample, we consider the encoder is very likely a trojaned encoder.

\section{Evaluation} \label{sec:evaluation}

We use the following research questions~(RQs) to evaluate \toolname:

{\bf RQ1}: How effective is our method?

{\bf RQ2}: How efficient is our method?

{\bf RQ3}: How robust is our method against adaptive attack?

{\bf RQ4}: How effective is our method if the defender has no access to the pre-training dataset?

\subsection{Experiment Setup}

We employ five commonly used datasets, CIFAR10~\cite{krizhevsky2009cifar10learning}, GTSRB~\cite{Houben-IJCNN-2013}, SVHN~\cite{netzer2011reading}, STL-10~\cite{coates2011analysis}, and ImageNet~\cite{ILSVRC15}, for pre-training encoders and training downstream classifiers. We use three well-known model architectures, ResNet18, ResNet34, and ResNet50~\cite{he2016deep}. As the CLIP dataset~\cite{Radford2021clip} is not publicly available, we downloaded a pre-trained encoder from~\cite{openaiclip} and use ImageNet to finetune the encoder by applying SimCLR~\cite{chen2020simclr} algorithm.

For backdoor attacks, we consider three categories in the SSL setting, namely {\it Image-on-Image}, {\it Image-on-Pair}, and {\it Text-on-Pair}, as discussed in Section~\ref{sec:related_backdoor_self}. Note that there are only a limited number of public backdoored encoders, we hence use the official implementation~\cite{jia2022badencoder} or implement the attacks strictly following the original paper~\cite{carlini2022poisoning} to construct backdoored encoders. 
For {\it Image-on-Image} and {\it Image-on-Pair} attacks, we choose a ``priority'' image from GTSRB, a ``one'' image from SVHN, and a ``truck'' image from STL-10 as attack targets. We only consider backdoored encoders that achieve at least 99\% attack success rate in the targeted downstream classifiers. 
For {\it Text-on-Pair} attack, we choose the label text ``priority'' for GTSRB, ``one'' for SVHN, and ``truck'' for STL-10 to fill in a prompt list (shown in Table~\ref{tab:app:prompt_list} in Appendix~\ref{sec:app:attack_setting}) and use these text captions as attack targets. The $z$-score introduced in~\cite{carlini2022poisoning} quantifies to what extent the subject encoder is trojaned. We only consider backdoored encoders with a $z$-score greater than 2.5 for evaluation. 
We set $\beta=-0.99$ and $\tau=0.1$ during the detection.
We use 444 encoders (111 benign and 333 backdoored) to evaluate \toolname. Details are shown in Appendix~\ref{sec:app:eval_setup}.\looseness=-1

\subsection{RQ1: Effectiveness of Our Method}

We evaluate the performance of \toolname by using common metrics (e.g., detection accuracy, ROC-AUC). We also show the distributions of inverted triggers for clean and backdoored encoders and study how the two sets are separated by \toolname.\looseness=-1

\begin{table*}[t]
  \centering
  \footnotesize
  \tabcolsep=5.6pt
  \small\addtolength{\tabcolsep}{-3pt}
  \caption{Detection Performance. The first three columns list the attack category, the dataset used for pre-training encoders, and the model architecture. RN18, RN34, and RN50 denote model architecture ResNet18, ResNet34, and ResNet50, respectively. The following three column blocks present the results for which dataset the attack target comes from, i.e., GTSRB, SVHN, and STL-10. 
  Columns in each block show the number of true positives (TP), false positives (FP), false negatives (FN), true negatives (TN) when we use \toolname to detect backdoored encoders. Acc denotes the overall detection accuracy.
}
  \label{tab:detect-performance}
  \begin{tabular}{cccccccccccccccccc}
    \toprule
    \multirow{2}{*}{\makecell[c]{Attack\\ Category}}  & 
    \multirow{2}{*}{\makecell[c]{Pre-training \\ Dataset}} &
    \multirow{2}{*}{\makecell[c]{Model \\ Arch}} &
    \multicolumn{5}{c}{GTSRB atk} &
    \multicolumn{5}{c}{SVHN atk} &
    \multicolumn{5}{c}{STL-10 atk}\\
    \cmidrule(lr){4-8} \cmidrule(lr){9-13}\cmidrule(lr){14-18}  
    ~ & ~ & ~ & TP & FP & FN & TN & Acc & TP & FP & FN & TN  & Acc & TP & FP & FN & TN & Acc\\
    \midrule
    \multirow{5}{*}{\makecell[c]{Img-on-Img}} & \multirow{3}{*}{\makecell[c]{CIFAR10}} &
    RN18 & 30 & 2 & 0 & 28 & 96.7 & 30 & 2 & 0 & 28 & 96.7 & 30 & 2 & 0 & 28 & 96.7 \\
    ~ & ~ & RN34 & 30 & 0 & 0 & 30 & 100  & 30 & 0 & 0  & 30 & 100 & 30 & 0 & 0 & 30 & 100 \\
    ~ & ~ & RN50 & 15 & 0 & 0 & 15 & 100  & 15 & 0 & 0 & 15 & 100  & 15 & 0 & 0 & 15 & 100\\
    \cmidrule(rl){2-3}
    ~ & ImageNet & RN50  & 12 & 0 & 0 & 12 & 100  & 12 & 0 & 0 & 12 & 100  & 12 & 0 & 0 & 12 & 100\\
    \midrule
    Img-on-Pair & CLIP & RN50 & 12 & 0 & 0 & 12 & 100  & 12 & 0 & 0 & 12 & 100  & 12 & 0 & 0 & 12 & 100\\
    \midrule
    Text-on-Pair & CLIP & RN50 & 12 & 0 & 0 & 12 & 100  & 9 & 0 & 3 & 12 & 87.5  & 12 & 0 & 0 & 12 & 100\\
    \midrule
    Summary & - & - & 111 & 2 & 0 & 109 & {\bf 99.1} & 108 & 2 & 3 & 109 & {\bf 97.7} & 111 & 2 & 0 & 109 & {\bf 99.1}\\
    \bottomrule
  \end{tabular}
  \vspace{-5pt}
\end{table*}

\begin{figure}
    
    \centering
    \begin{subfigure}[t]{.48\linewidth}
        \centering
        \includegraphics[width=\textwidth]{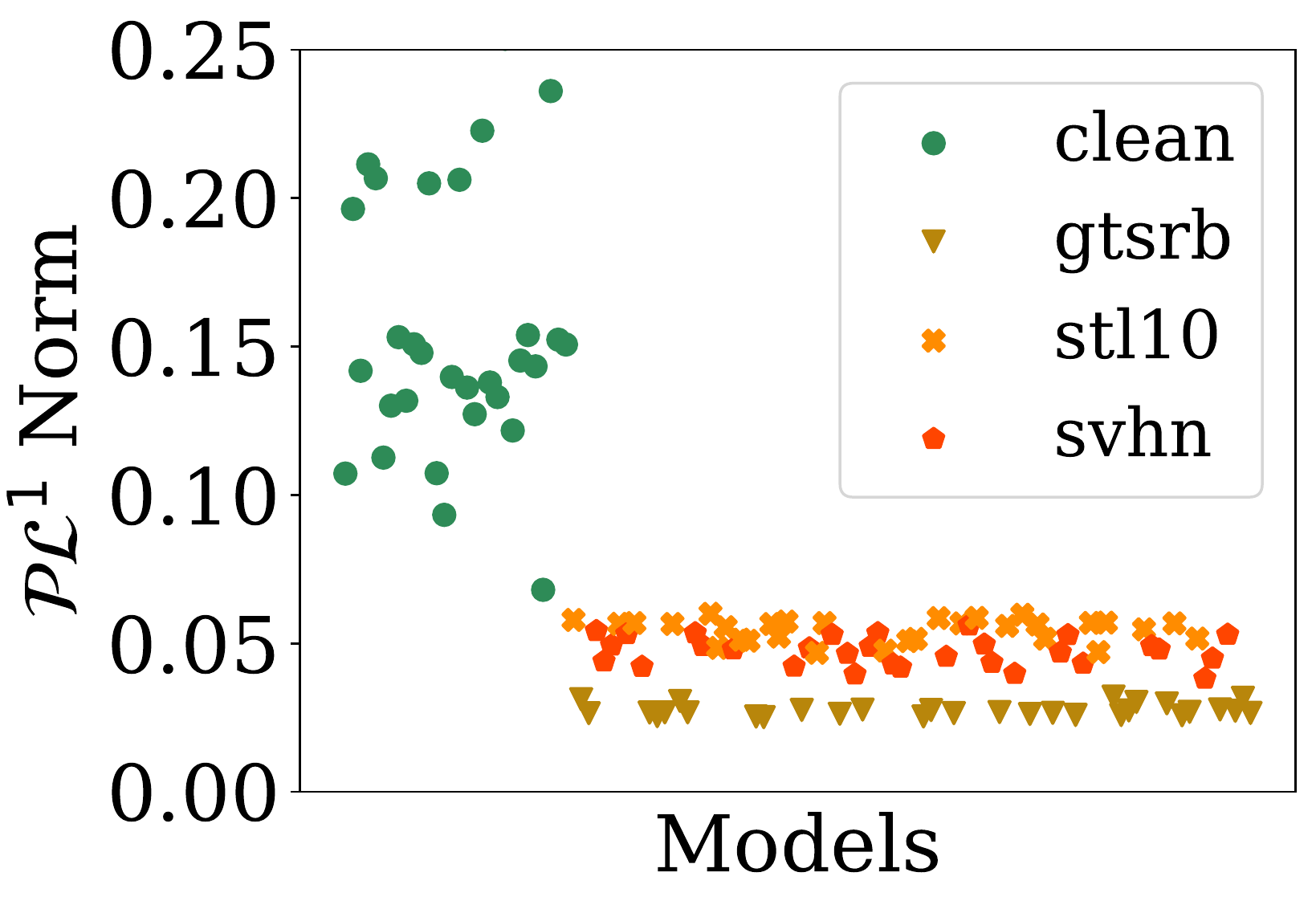}
        \vspace{-15pt}
        \caption{CIFAR10-ResNet18}
        \label{fig:c10rn18}
    \end{subfigure}
    \begin{subfigure}[t]{.48\linewidth}
        \centering
        \includegraphics[width=\textwidth]{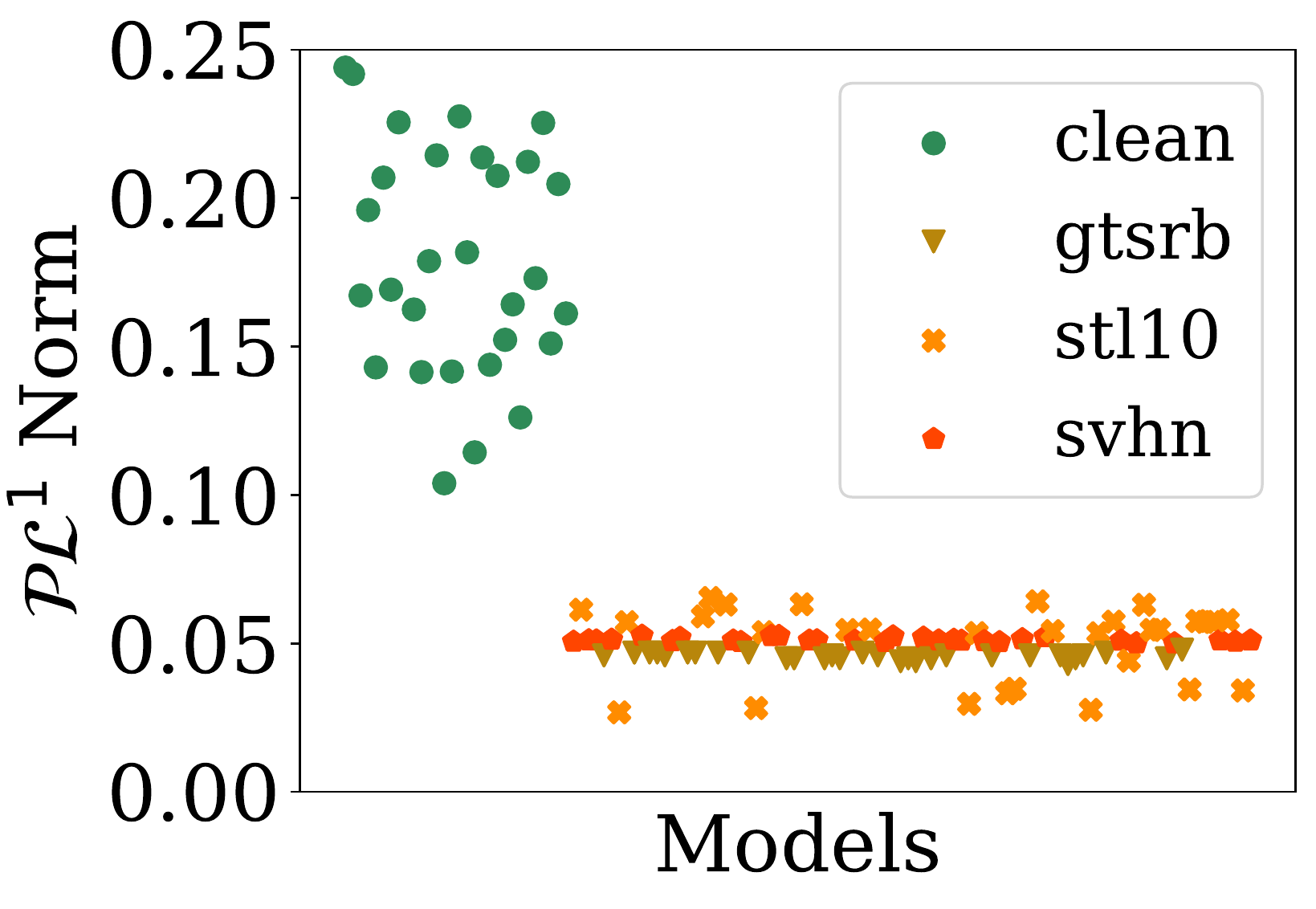}
        \vspace{-15pt}
        \caption{CIFAR10-ResNet34}
        \label{fig:c10rn34}
    \end{subfigure}
    \begin{subfigure}[t]{.48\linewidth}
        \centering
        \includegraphics[width=\textwidth]{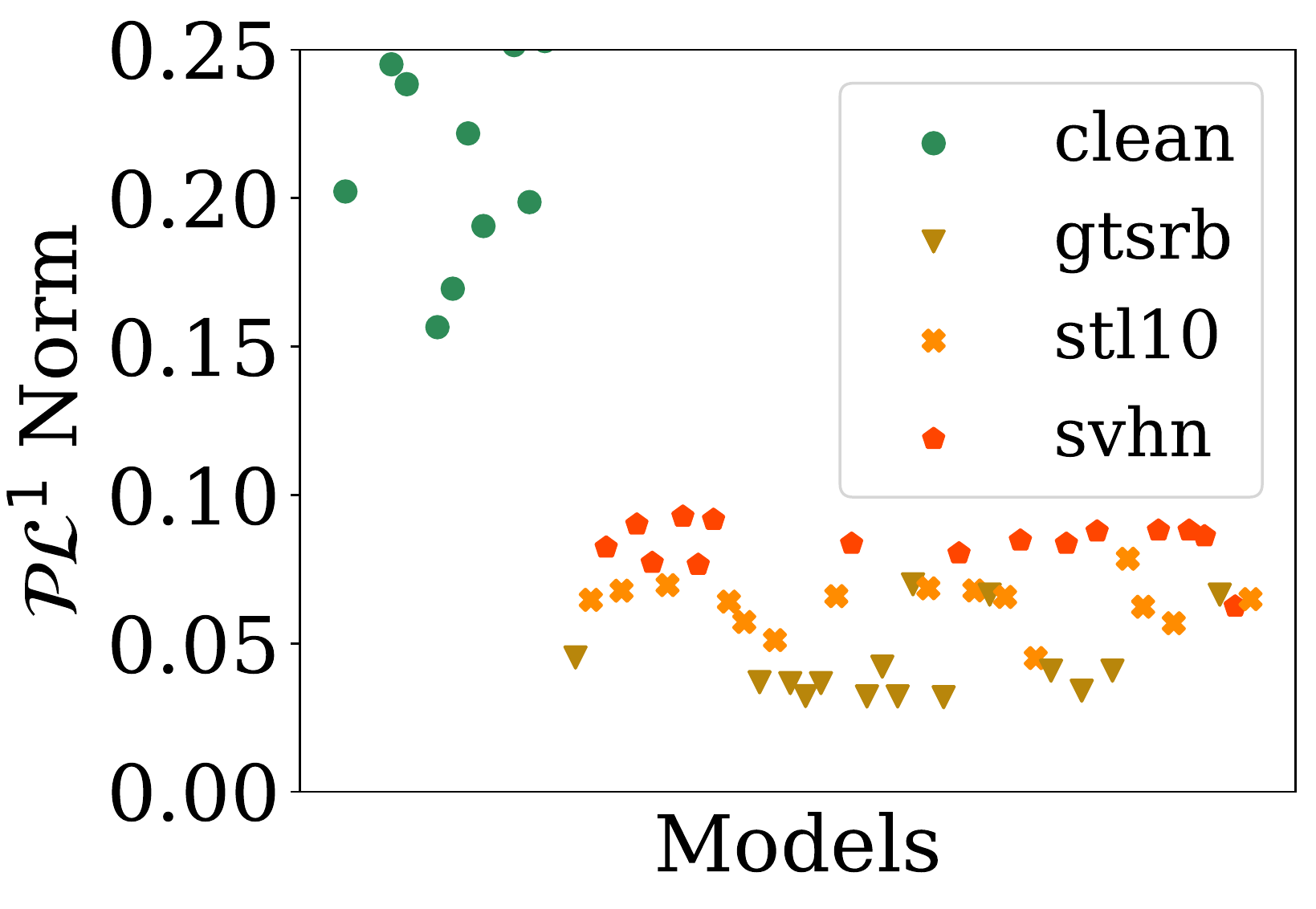}
        \vspace{-15pt}
        \caption{CIFAR10-ResNet50}
        \label{fig:c10rn50}
    \end{subfigure}
    \begin{subfigure}[t]{.48\linewidth}
        \centering
        \includegraphics[width=\textwidth]{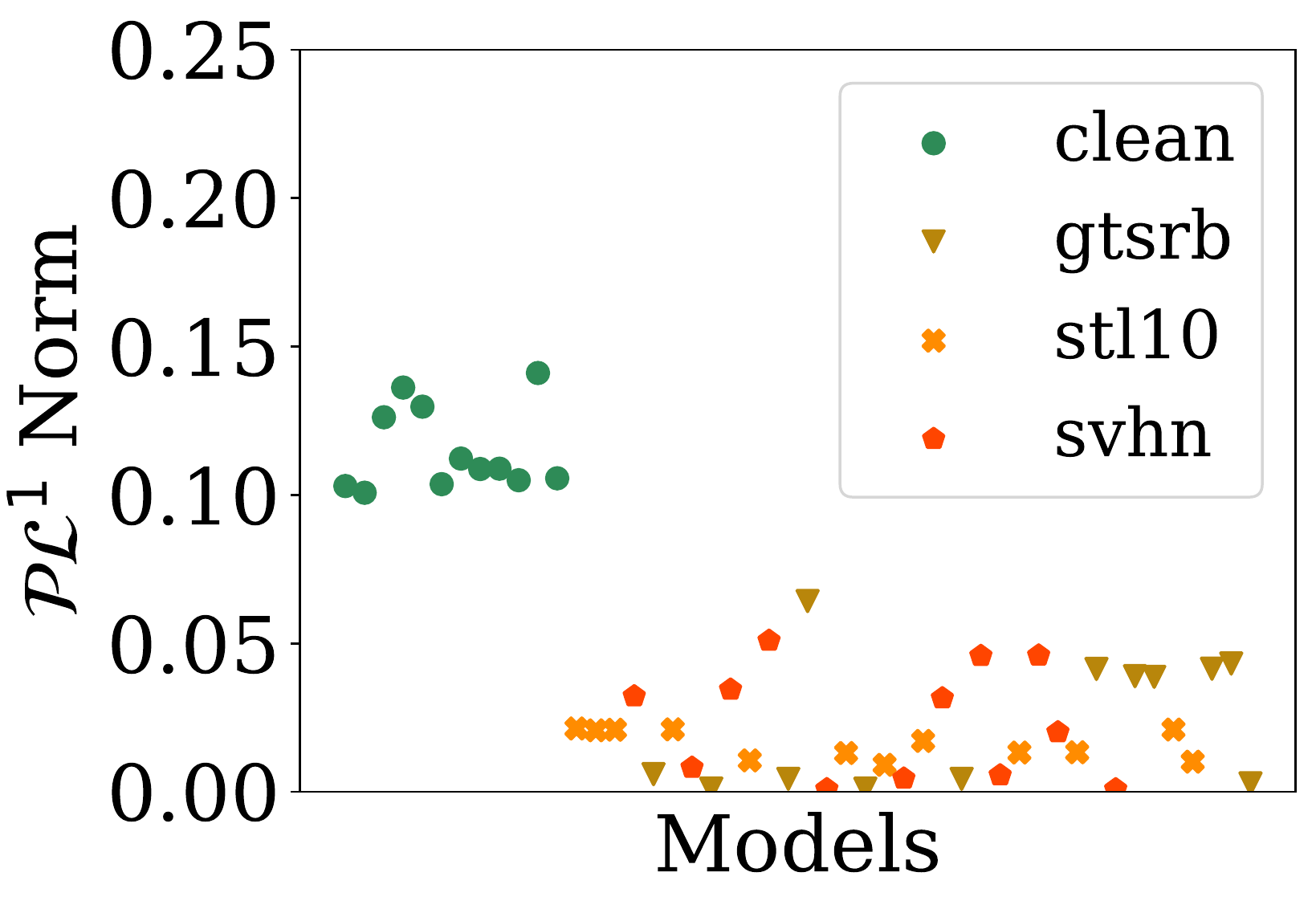}
        \vspace{-15pt}
        \caption{ImageNet-Resnet50}
        \label{fig:imgnet-rn50}
    \end{subfigure}
    \begin{subfigure}[t]{.48\linewidth}
        \centering
        \includegraphics[width=\textwidth]{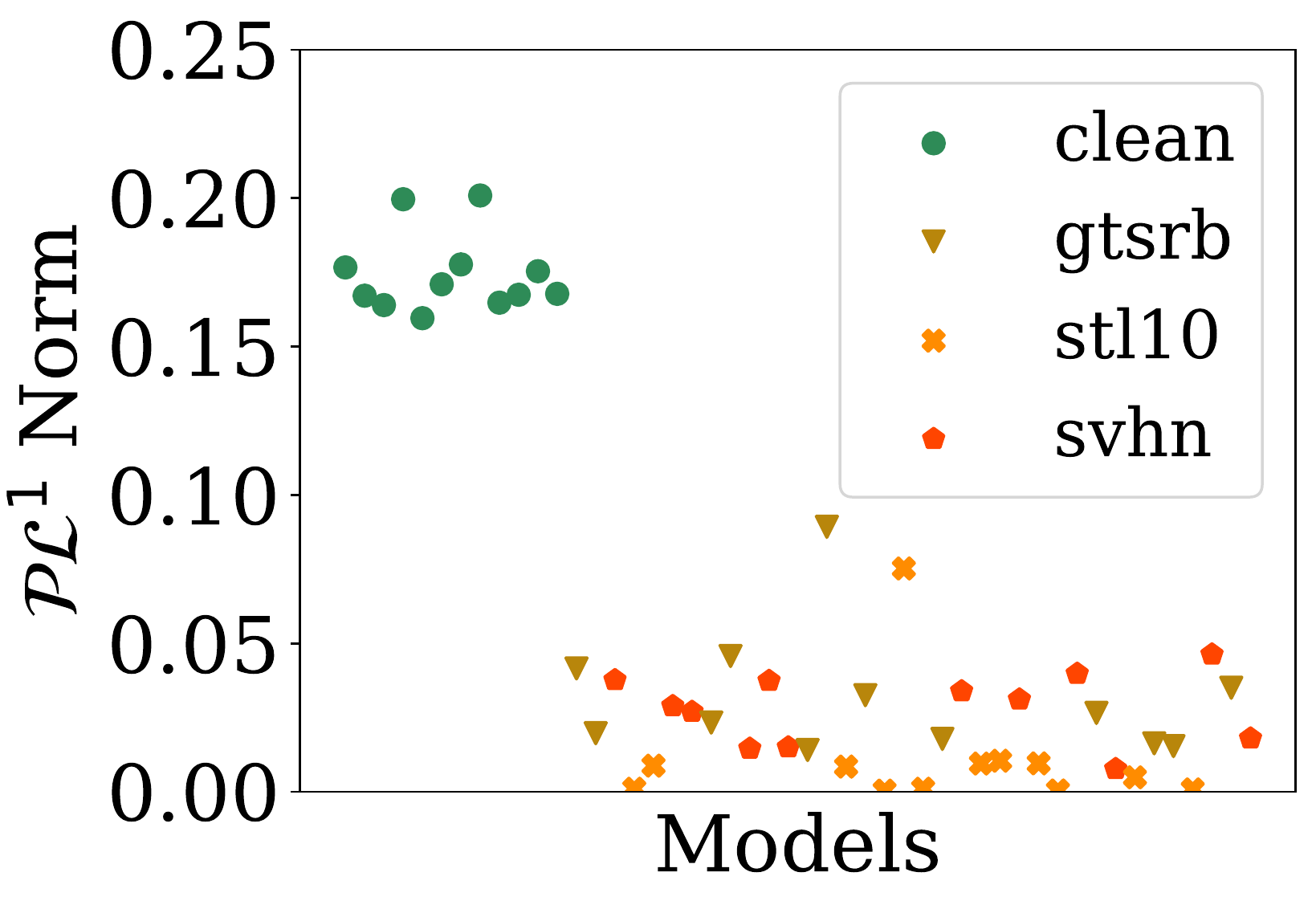}
        \vspace{-15pt}
        \caption{CLIP-Image-ResNet50}
        \label{fig:clip-img}
    \end{subfigure}
    \begin{subfigure}[t]{.48\linewidth}
        \centering
        \includegraphics[width=\textwidth]{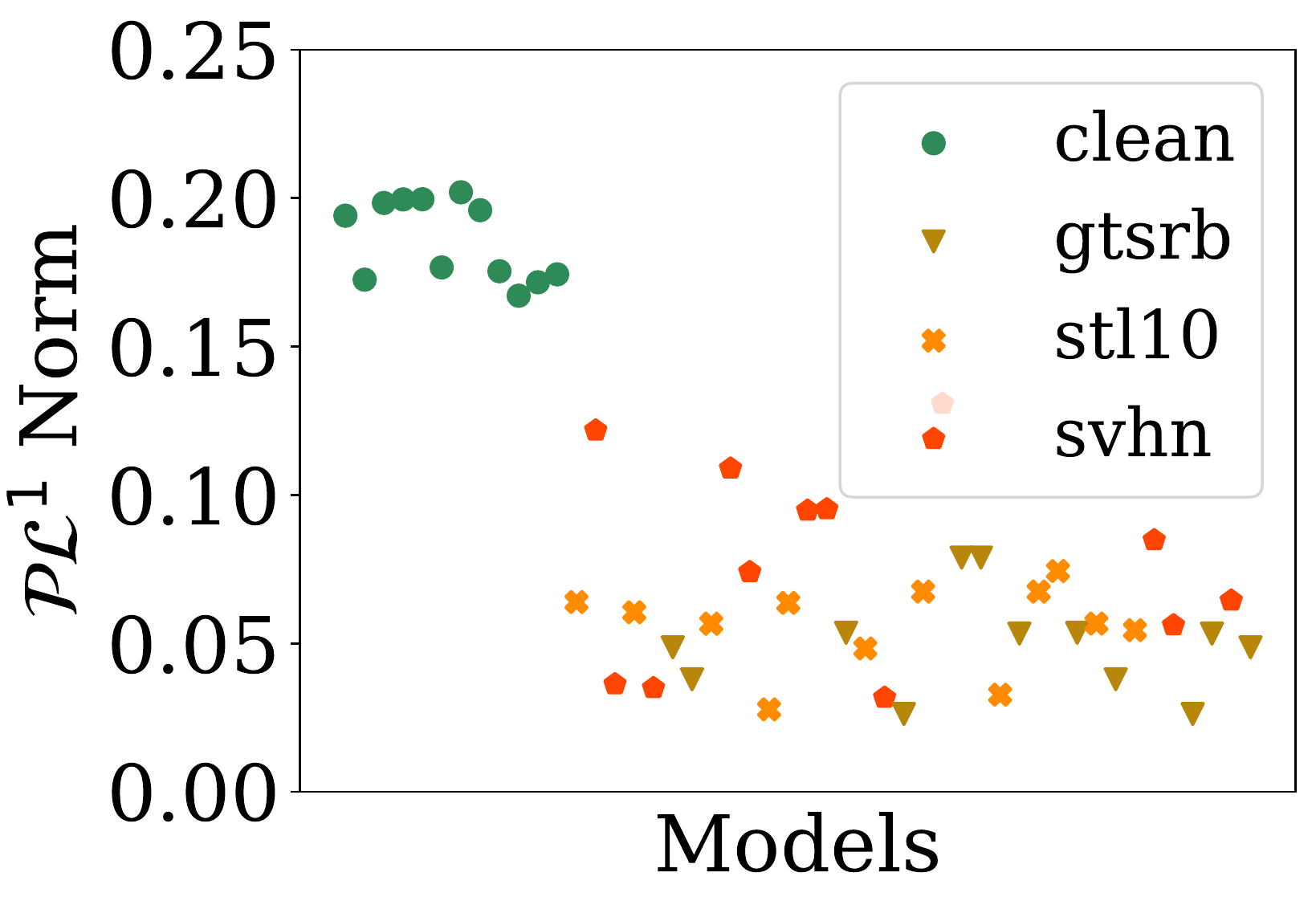}
        \vspace{-15pt}
        \caption{CLIP-Text-ResNet50}
        \label{fig:clip-text}
    \end{subfigure}
    \caption{Distribution of Inverted Triggers. 
    Each sub-figure corresponds to one setting (one line in Table~\ref{tab:detect-performance}) and depicts the results for that setting.  
    The $\vx$-axis denotes different models and the $\vy$-axis denotes the \metricname value. The green markers denote inverted triggers for clean encoders while other color markers (i.e., brown, orange, and red markers) denote inverted triggers for backdoored encoders with attack targets coming from GTSRB, STL-10 and SVHN, respectively.\looseness=-1
    }
    \label{fig:dist-trigger}
    \vspace{-10pt}
\end{figure}

The detection results of \toolname are shown in Table~\ref{tab:detect-performance}. 
We evaluate on three attack categories, namely {\it Image-on-Image}, {\it Image-on-Pair}, and {\it Text-on-Pair}.
For each attack category, we choose three attack targets, from GTSRB, SVHN and STL-10 respectively.

Observe that \toolname can effectively detect almost all the backdoored encoders with more than 95\% accuracy in most cases. Particularly, for 14 out of 18 scenarios, \toolname has 100\% detection accuracy.
For {\it Text-on-Pair} on SVHN, the detection accuracy is slightly lower (87.5\%). This is because the attack targets for this case are natural language sentences, and they usually have multiple target instances. For example, a trigger with the label text ``truck'' can use both ``a picture of truck'' and ``a nice photo of truck'' as attack targets, making the triggers less centralized than those attacks on images. Note that we use the same threshold for all the application/attack settings. That said, with the knowledge of the particular application scenario ({\it Text-on-Pair}), \toolname can still effectively distinguish backdoored encoders from clean encoders by slightly increasing the threshold, as depicted in Figure~\ref{fig:clip-text}.
The last row in Table~\ref{tab:detect-performance} show the summarized performance. We can see that \toolname achieves a detection accuracy of near 100\% in all cases on average, delineating its effectiveness. We also use the ROC (Receiver Operating Characteristic) curve to study the relation between true positive rate and false positive rate as shown in Figure~\ref{fig:roc} in Appendix~\ref{sec:app:roc}.

We study the distributions of inverted triggers for clean and backdoored encoders, which are shown in Figure~\ref{fig:dist-trigger}. Each sub-figure corresponds to one setting (one line in Table~\ref{tab:detect-performance}) and depicts the results for that setting. 
Observe that in all scenarios, inverted triggers for backdoored encoders have smaller \metricname than those for clean encoders. The triggers for backdoored encoders tend to cluster in small \metricname values ($< 0.1$). This demonstrates the reason why \toolname is able to effectively detect backdoored encoders with a same threshold.
We also visually show that the inverted triggers for backdoored encoders have much fewer perturbed pixels compared to those for clean encoders. Please see detailed results and discussion in Appendix~\ref{sec:app:trigger}.

\subsection{RQ2: Efficiency of Our Method}

\begin{figure}[t!]
    \begin{minipage}[c]{0.48\linewidth}
        \vspace{-15pt}
        \centering
        \includegraphics[width=\textwidth]{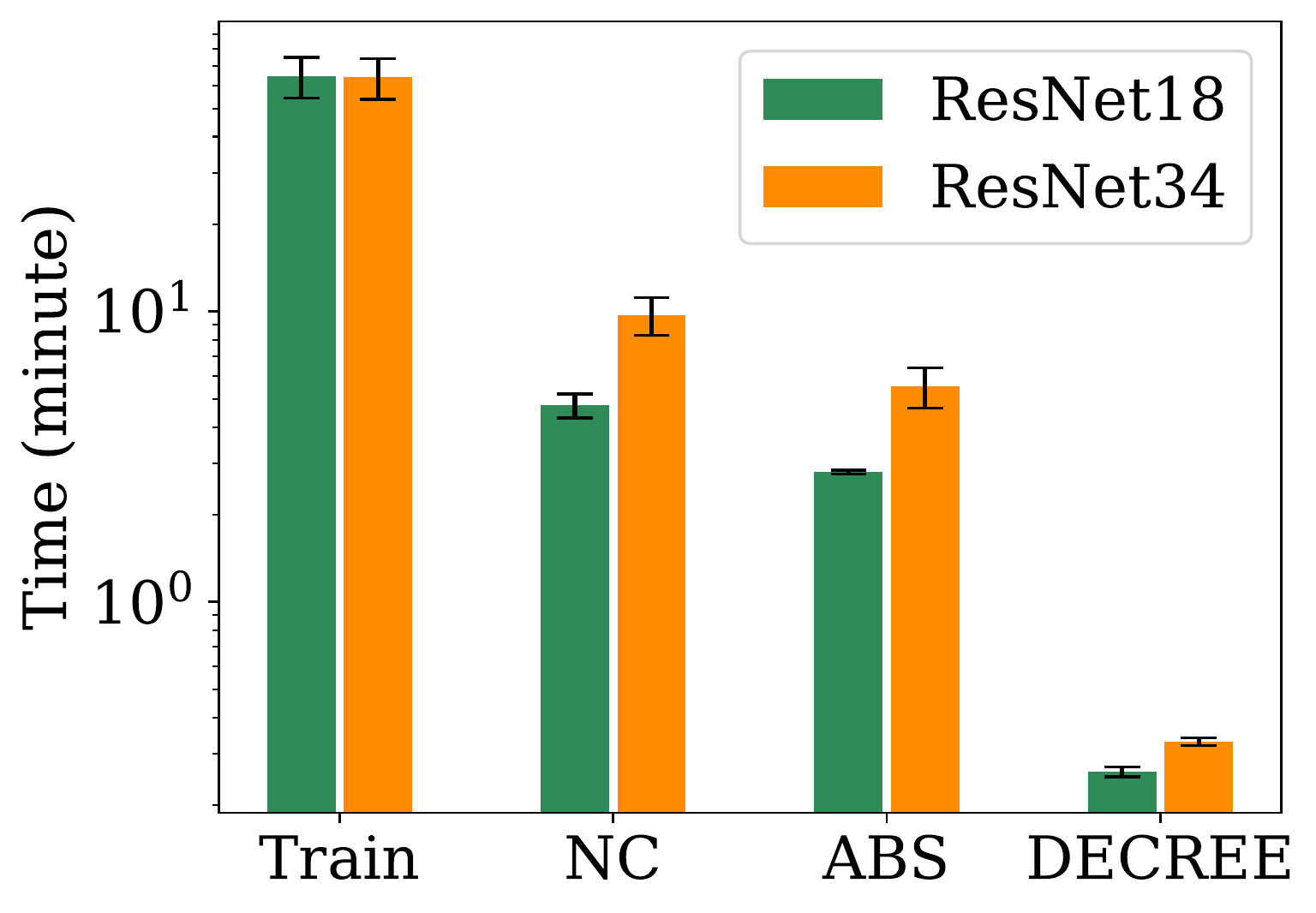}
        \caption{Time Efficiency}
        \label{fig:run-time}
        \vspace{-15pt}
    \end{minipage}
    ~
    \begin{minipage}[c]{0.48\linewidth}
        \vspace{-1pt}
        \centering
        \includegraphics[width=\textwidth]{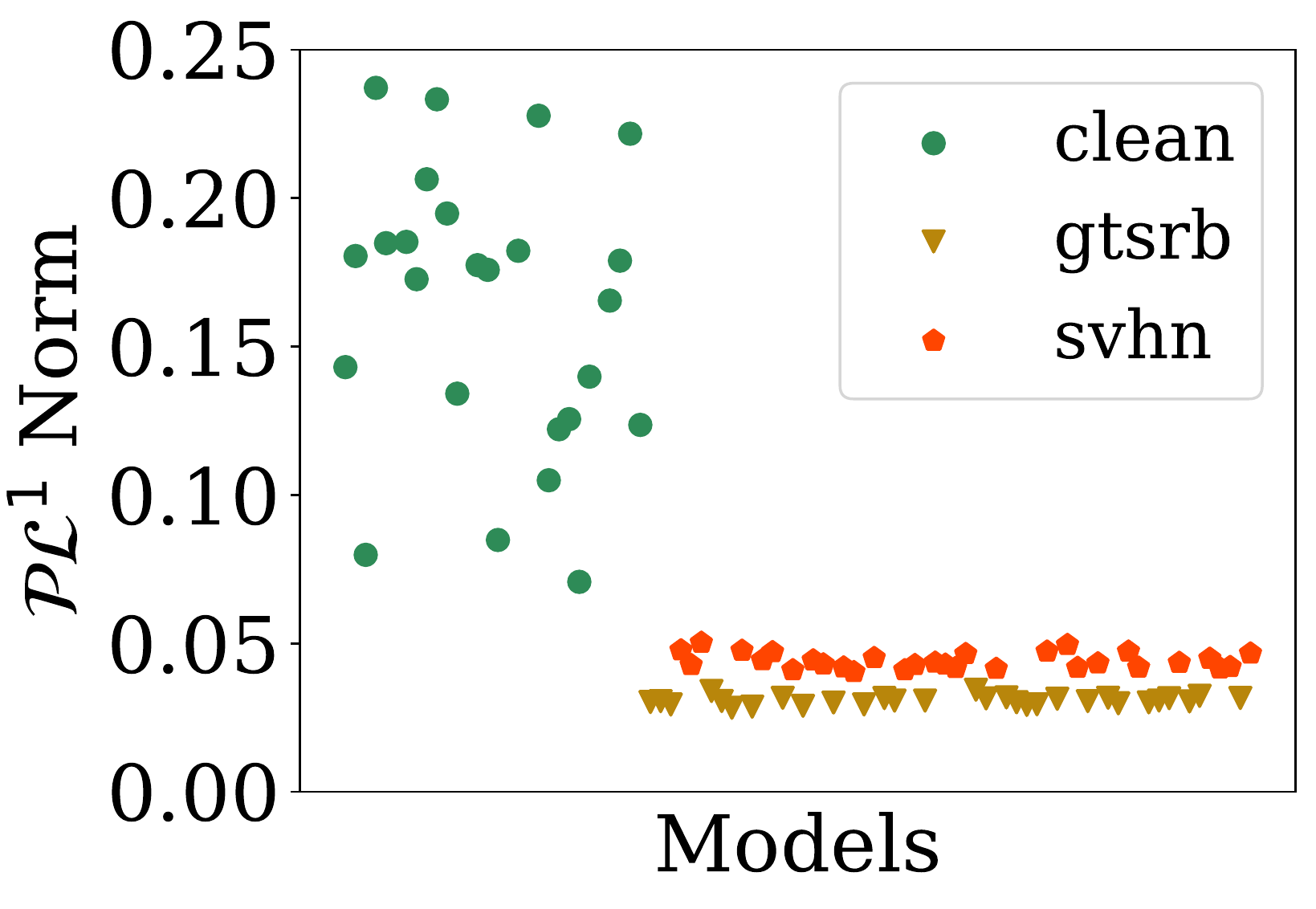}
        \caption{Performance without Access to Pre-training Dataset}
        \label{fig:no-acc}
        \vspace{-5pt}
    \end{minipage}
    \vspace{-10pt}
\end{figure}

In this section, we evaluate the efficiency of \toolname in comparison with two SOTA backdoor scanning techniques, i.e., Neural Cleanse (NC)~\cite{wang2019neural} and ABS~\cite{liu2019abs}. Recall in Section~\ref{sec:motivation}, we observe that existing detection methods need the knowledge of downstream tasks. In addition, they also require samples from the downstream dataset for detection. For a fair comparison, we assume existing detectors have full access to the downstream dataset, with which they can train a corresponding downstream classifier and perform the detection based on the classifier and downstream task samples.\looseness=-1

We conduct experiments on 10 backdoored encoders trained on CIFAR-10 with ResNet18 and ResNet34 architectures. The attack target is a ``one'' image from the SVHN dataset.
Figure~\ref{fig:run-time} shows the results. 
As existing techniques need to train downstream classifiers, we also show the training time of classifiers in the first two columns.\looseness=-1

Observe that training a classifier takes a large amount of time, more than 1 hour. The runtime of existing techniques is around 2-10 minutes. \toolname, on the other hand, only takes 15-20 seconds. It is 6-30 times faster than baselines, even without considering the training time for downstream classifiers. This is because DECREE
generates just one trigger for each encoder and do
not have to scan each label like what existing methods do.

\subsection{RQ3: Adaptive Attack}
\label{sec:eval:adaptive_attack}
We consider a stronger attack that aims to evade the detection of \toolname{} with the full knowledge of our detection pipeline.
Assume the loss function used in the original attack is $\gL_{atk}$. The stronger attack also considers $\gL_{sim}$, the same as $\gL$ in Eq.~\ref{eq:avg_sim}.
$\gL_{sim}$ quantifies the similarity among inputs stamped with the trigger. The attacker aims to enlarge this loss to make those samples less similar.
Therefore, the objective of the adaptive attack is as follows.

\vspace{-10pt}
\begin{equation}
  \argmin\limits \gL_{adapt} = \gL_{atk} -
  \alpha \cdot \gL_{sim}, \quad \alpha>0
\label{eq:adapt_attack_loss}
\end{equation}
We conduct experiments on a ResNet18 encoder trained on CIFAR10 and the attack target is a “one” image from the
SVHN dataset. We set $\alpha=1$. The adaptive attack can produce a trojaned encoder that has an inverted trigger with a \metricname of 0.14
, evading \toolname's detection. However, the ASR on downstream STL-10 degrades from 99.9\% to 69.9\%. Intuitively, $\gL_{adapt}$ forces the embeddings of inputs stamped with the trigger to have a similar embedding with the attack target while trying to make them orthogonal to each other. It hence is difficult for the attack to achieve a high ASR and evade our detection (i.e. inputs stamped with triggers share high cosine similarity) at the same time. Please see more details in Appendix~\ref{sec:app:adaptive_attack}.

\subsection{RQ4: No Access to Pre-training Dataset}\label{sec:eval_no_access}

In previous experiments, we use a small subset of the pre-training dataset for trigger inversion. In extreme cases, the pre-training dataset may not be available, which significantly increases the difficulty of backdoor scanning. We evaluate \toolname in this setting to show its robustness.
We use CIFAR10 as the pre-training dataset, GTSRB and SVHN as origins of attack targets, and STL-10 as the shadow dataset for detection. As shown in Figure~\ref{fig:no-acc}, the distribution of inverted triggers in this setting is similar to those in Figure~\ref{fig:dist-trigger}.
\toolname can clearly separate clean and backdoored encoders based on \metricname, delineating the generalizability of \toolname.
Note that CLIP pre-training dataset is not public. Rows {\it Image-on-Pair} and {\it Text-on-Pair} in Table~\ref{tab:detect-performance} also fall into this challenging threat model. Figure~\ref{fig:clip-img} and Figure~\ref{fig:clip-text} show the detection results for these two. 

One key factor contributing to the generalizability of \toolname is that encoders pre-trained on unlabeled data via contrastive learning do not easily overfit on a certain dataset. In addition, \metricname considers different input dimensions so that \toolname is insensitive to different attack settings.

\subsection{Other Experiments}

\noindent \textbf{Ablation Study.} 
We conduct ablation studies to validate the robustness of \toolname against various trigger configurations (e.g., color, size, texture) and different attack strategies. 
\camera{Details are shown in Appendix~\ref{sec:app:different_triggers}.}
We also study the hyper-parameters 
\camera{(shadow dataset size $M$ and decision threshold $\tau$)}
and show the performance is insensitive to different hyper-parameters.
Details can be found in Appendix~\ref{sec:app:hyper_params}.

\camera{
\noindent \textbf{Advanced Attacks.} 
We adapt 2 dynamic attacks~\cite{nguyen2020wanet,Li2021invisible} from supervised learning into our settings
and find that such attacks can hardly succeed in SSL setting. Details can be found in Appendix~\ref{sec:app:advanced_attack}.
}

\camera{
\noindent \textbf{More SSL Attacks.} 
We also study 3 emerging attacks~\cite{Saha2022backdoorSSL, li2022demystifying, zhang2023corruptencoder}. We find that \toolname can detect acute attacks (i.e., high ASR) with patch-like triggers~\cite{zhang2023corruptencoder}, but may fail on attacks with pervasive triggers~\cite{li2022demystifying} or stealthy attacks~\cite{Saha2022backdoorSSL}.
Details can be found in Appendix~\ref{sec:app:ssl_attack}.
}

\section{Conclusion} \label{sec:conclusion}
We propose the first backdoor detection method \toolname for pre-trained encoders. Our method fills in the gap where existing detection techniques only focus on supervised learning scenarios. Our evaluation shows that \toolname can effectively and efficiently separate benign and trojaned encoders. Our method is also robust against adaptive attacks and generalizes to a more challenging threat model.

\noindent
\textbf{Limitation of Our Work.} We currently do not handle text-format trigger. Our method mainly focuses on three types of attacks ({\it Image-on-Image}, {\it Image-on-Pair}, and {\it Text-on-Pair}), the attack subject of which is an image encoder. For {\it Text-on-Text} attack, it introduces extra challenges to invert text-format triggers as the input in NLP is discrete (e.g., words), different from the pixel values in computer vision.

\section*{Acknowledgement}
\camera{
We thank the anonymous reviewers for their constructive comments. This research was supported, in part by IARPA TrojAI W911NF-19-S-0012, NSF 1901242 and 1910300, ONR N000141712045, N000141410468 and N000141712947. Any opinions, findings, and conclusions in this paper are those of the authors only and do not necessarily reflect the views of our sponsors.}

{\small
\bibliographystyle{ieee_fullname}
\bibliography{egbib}
}

\clearpage

\appendix

\section*{Appendix} \label{sec:appendix}

\noindent We provide a table of contents below for better navigation of the appendix.

\noindent {\bf Appendix~\ref{sec:app:eval_setup}} provides details of evaluation setup.

\noindent {\bf Appendix~\ref{sec:app:attack_setting}} introduces the settings of backdoor attacks on self-supervised learning that are adopted in our evaluation.

\noindent {\bf Appendix~\ref{sec:app:trigger}} studies the triggers inverted by \toolname.

\noindent {\bf Appendix~\ref{sec:app:roc}} uses ROC curve to quantify the effectiveness of \toolname.

\noindent {\bf Appendix~\ref{sec:app:efficiency}} evaluates the efficiency of \toolname in comparison with two SOTA backdoor scanning techniques.

\noindent {\bf Appendix~\ref{sec:app:adaptive_attack}} designs an adaptive attack aiming to evade our detection.

\noindent {\bf Appendix~\ref{sec:app:different_triggers}} studies the effectiveness of \toolname against different trigger patterns and sizes.

\noindent {\bf Appendix~\ref{sec:app:hyper_params}} shows the effectiveness of threshold $\tau$. \looseness=-1

\noindent {\bf Appendix~\ref{sec:app:advanced_attack}} explores the feasibility of adapting 2 existing advanced attacks from supervised learning into self-supervised learning setting.

\noindent {\bf Appendix~\ref{sec:app:ssl_attack}} discusses on 3 emerging SSL backdoor attacks.

\begin{table*}[ht!]
  \centering
  \footnotesize
    \small\addtolength{\tabcolsep}{-3pt}
  \caption{Model Statistics}
  \label{tab:model_statistics}
  \begin{tabular}{ccccccccccc}
    \toprule
    \multirow{2}{*}{\makecell[c]{Attack\\Category}} &
    \multirow{2}{*}{\makecell[c]{Pre-training \\ Dataset}} &
    \multirow{2}{*}{\makecell[c]{Model\\Arch}} &
    \multirow{2}{*}{\makecell[c]{Input\\Size}} &
    \multirow{2}{*}{\makecell[c]{\#Params}} &
    \multirow{2}{*}{\makecell[c]{Clean \\Encoder}} &
    \multicolumn{3}{c}{Attack Datasets} \\

    \cmidrule(lr){7-9} 
    ~& ~ & ~ & ~ & ~ & ~ & GTSRB & SVHN & STL-10
    \\
    \midrule
    \multirow{5}{*}{\makecell[c]{{\it Image-on-Image}}} & \multirow{3}{*}{\makecell[c]{CIFAR10}} &
    ResNet18 & 32$\times$32$\times$3 & 11,168,832 & 30 & 30 & 30 & 30 \\
    ~ & ~ & ResNet34 & 32$\times$32$\times$3 & 21,276,992 & 30 & 30 & 30 & 30  \\
    ~ & ~ & ResNet50 & 32$\times$32$\times$3 & 23,500,352 &15 & 15 & 15 & 15  \\
    \cmidrule(lr){2-3}
    ~ & ImageNet & ResNet50 &224$\times$224$\times$3 & 25,557,032 & 12 & 12 & 12 & 12 \\
    \midrule
    {\it Image-on-Pair} & CLIP Dataset & ResNet50 &224$\times$224$\times$3 & 38,316,896& 12 & 12 & 12 & 12\\
    \midrule
    {\it Text-on-Pair} & CLIP Dataset & ResNet50 &224$\times$224$\times$3 & 38,316,896 & 12 & 12 & 12 & 12 \\
    \bottomrule
  \end{tabular}
\end{table*}

\section{Evaluation Setup}\label{sec:app:eval_setup}

Table~\ref{tab:model_statistics} shows the statistics of evaluated attacks, datasets, and encoders. Column 1 denotes the attack category. Column 2 shows the pre-training datasets used for constructing encoders. Columns 3-5 present the model architecture, input image shape, and the number of (trainable) model parameters. Column 6 shows the number of clean encoders for each setting. For backdoored encoders, we choose one label from each {\it attack datasets} as attack target label. For example, when attack dataset is GTSRB, we choose a ``priority'' image as attack target in {\it Image-on-Image} and {\it Image-on-Pair} settings and choose the word ``priority'' to fill in prompts in {\it Text-on-Pair} setting. We introduce more details in Appendix~\ref{sec:app:attack_setting}. We evaluate on three attack datasets that are shown in Columns 7-9. The numbers denote how many backdoored encoders are trained for the corresponding attack datasets. In total, we have 444 encoders (111 benign and 333 backdoored). \looseness=-1

\section{Attack Settings}
\label{sec:app:attack_setting}

\subsection{Image-on-Image \& Image-on-Pair}
For {\it Image-on-Image} and {\it Image-on-Pair} attacks, we follow the code released by BadEncoder~\cite{jia2022badencoder} to construct backdoored encoders. Specifically, the main idea is that, given a clean encoder $E$, the attacker aims to get a trojaned encoder $E'$ such that $E$ and $E'$ satisfy the following 3 properties: (1) For each clean input image $x$, $E(x)$ and $E'(x)$ should be similar. (2) For the target image $r$, $E(r)$ and $E'(r)$ should be similar. (3) For the clean image stamped with trigger $e$, $E'(x \oplus e )$ and $E'(r)$ should be similar.

For each attack datasets, we use the same target images as ~\cite{jia2022badencoder}. We select trojaned encoders that can train downstream classifiers with ASR $>$ 99\% and accuracy $>$ 70\%. \looseness=-1

\subsection{Text-on-Pair}

For {\it Text-on-Pair} attack, we follow the method introduced in ~\cite{carlini2022poisoning}. The main idea is to construct a malicious training dataset $\gP$ (size of which is a small fraction of pre-training dataset size). $\gP$ is defined as $\gP=\{(x_i \oplus e, c)\}_i$, where $x_i$ are clean images, $e$ is trigger and $c$ is attack target caption. The caption is formed by filling in prompts (shown in Table~\ref{tab:app:prompt_list}) with a word of interest from attack datasets(shown in Table~\ref{tab:app:target_word}). We choose backdoored encoders with $z$-score~\cite{carlini2022poisoning} higher than 2.5.

\begin{table}[h!]
  \centering
  \footnotesize
  \vspace{-10pt}
  \caption{Attack Target Words in {\it Text-on-Pair} Attack}
  \label{tab:app:target_word}
  \begin{tabular}{ll}
    \toprule
    Attack Dataset & Target Word \\
    \midrule
    GTSRB & ``priority'' \\
    SVHN & ``one'' \\
    STL-10 & ``truck'' \\
    \bottomrule
  \end{tabular}
\end{table}

\begin{table}[h!]
  \centering
  \footnotesize
  \caption{Prompt List in {\it Text-on-Pair} Attack}
  \label{tab:app:prompt_list}
  \begin{tabular}{l|l}
    \toprule
    ``a photo of a \{\}.'' & ``a photo of the \{\}.'' \\
    ``a blurry photo of a \{\}.'' & ``a blurry photo of the \{\}.'' \\
    ``a black and white photo of a \{\}.'' & ``a black and white photo of the \{\}.'' \\
    ``a low contrast photo of a \{\}.'' & ``a low contrast photo of the \{\}.'' \\
    ``a high contrast photo of a \{\}.'' & ``a high contrast photo of the \{\}.'' \\
    ``a bad photo of a \{\}.'' & ``a bad photo of the \{\}.'' \\
    ``a good photo of a \{\}.'' & ``a good photo of the \{\}.'' \\
    ``a photo of a small \{\}.'' & ``a photo of the  small \{\}.'' \\
    ``a photo of a big \{\}.'' & ``a photo of the big \{\}.'' \\
    \bottomrule
  \end{tabular}
\end{table}

\section{Triggers Inverted by \toolname}
\label{sec:app:trigger}
\begin{figure*}[h]
\centering
\begin{subfigure}[h]{\textwidth}
    \begin{minipage}[b]{.33\linewidth}
        \centering
        \includegraphics[width=\textwidth]{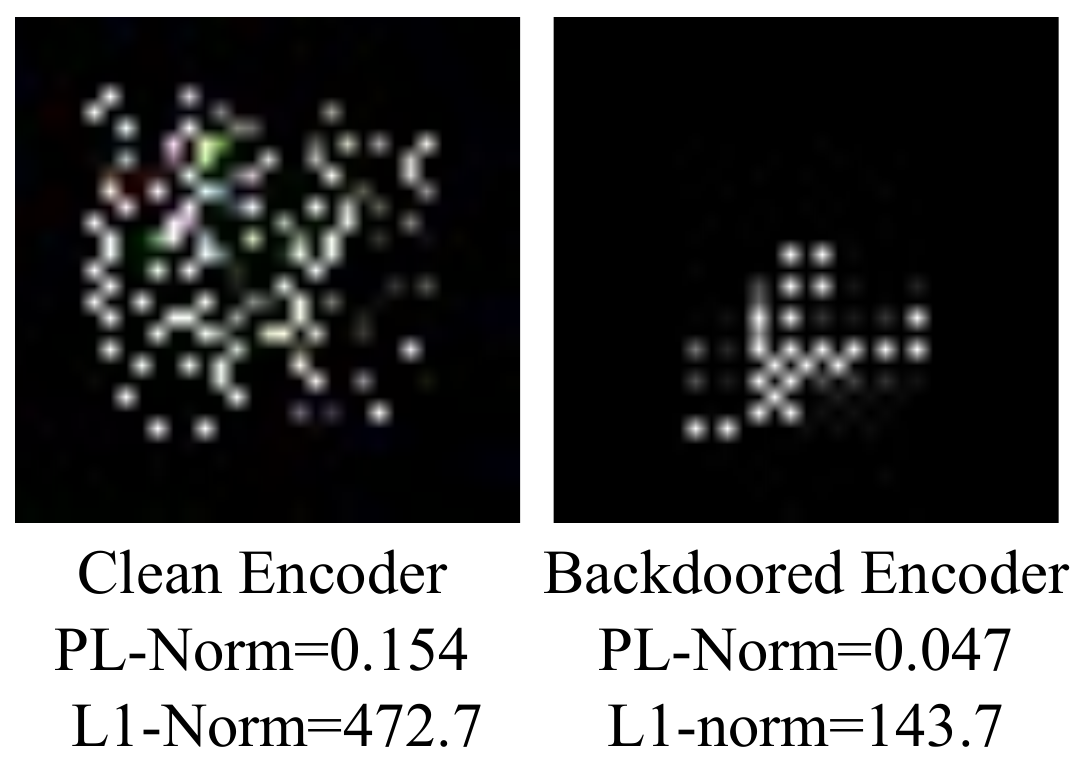}
        \caption{CIFAR10-ResNet18}
        \label{fig:inverted_cifar10_rn18}
    \end{minipage}
    \begin{minipage}[b]{.33\linewidth}
        \centering
        \includegraphics[width=\textwidth]{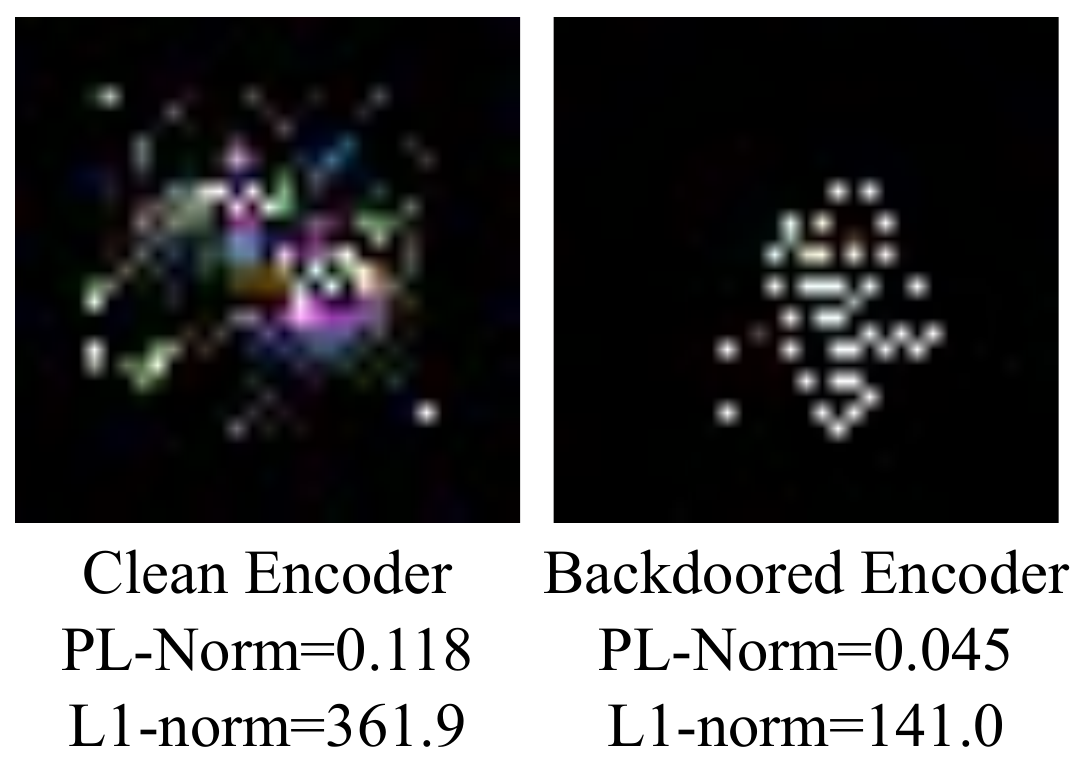}
        \caption{CIFAR10-ResNet34}
        \label{fig:inverted_cifar10_rn34}
    \end{minipage}
    \begin{minipage}[b]{.33\linewidth}
        \centering
        \includegraphics[width=\textwidth]{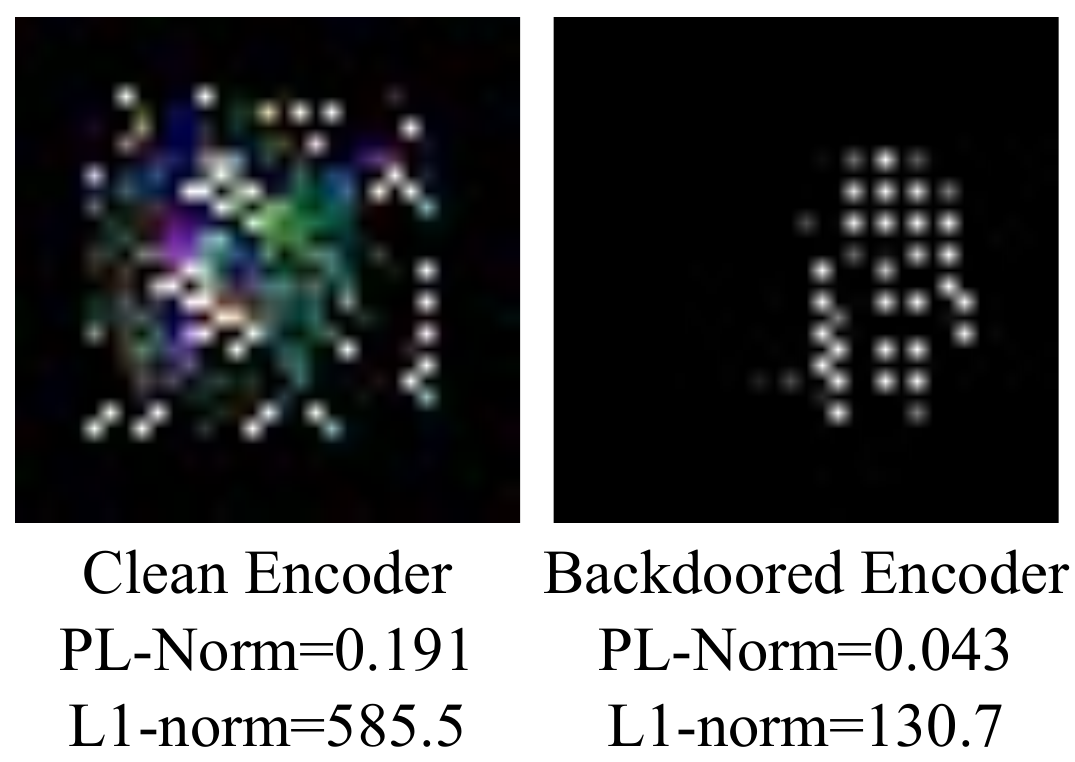}
        \caption{CIFAR10-ResNet50}
        \label{fig:inverted_cifar10_rn50}
    \end{minipage}
\end{subfigure}

\begin{subfigure}[h]{\textwidth}
    \begin{minipage}[b]{.33\linewidth}
        \centering
        \includegraphics[width=\textwidth]{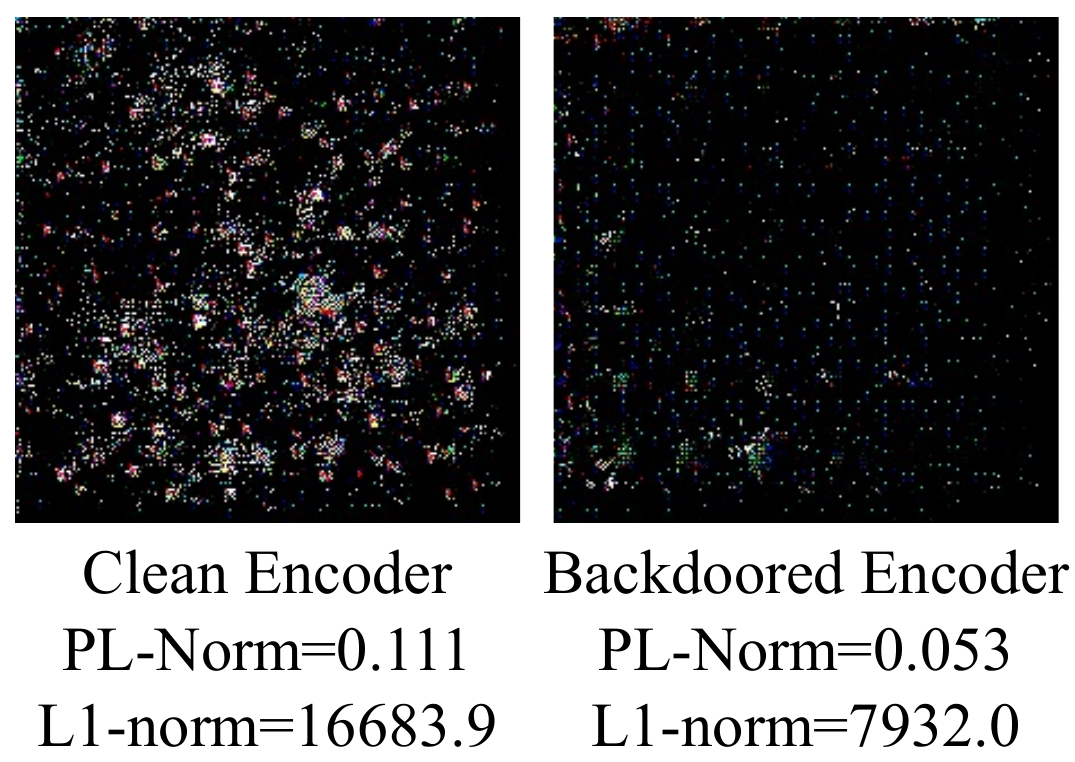}
        \caption{ImageNet-ResNet50}
        \label{fig:inverted_clip_text}
    \end{minipage}
    \begin{minipage}[b]{0.33\linewidth}
        \centering
        \includegraphics[width=\textwidth]{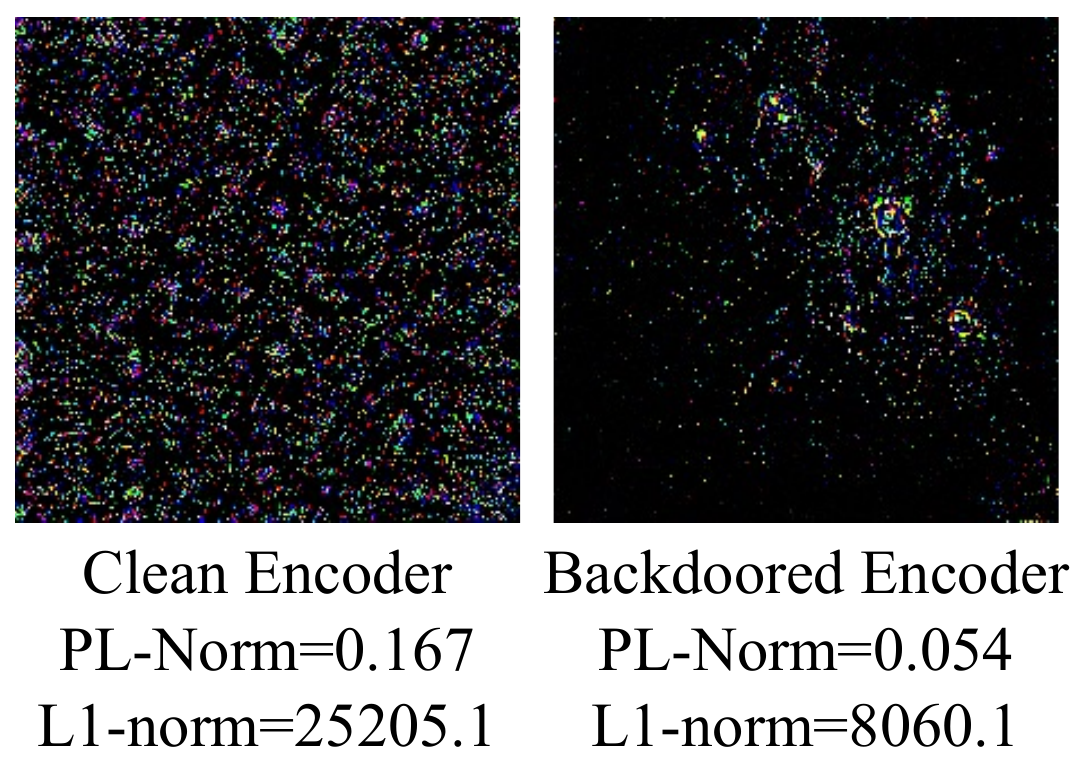}
        \caption{CLIP-Image}
        \label{fig:inverted_clip_img}
    \end{minipage}
    \begin{minipage}[b]{0.33\linewidth}
        \centering
        \includegraphics[width=\textwidth]{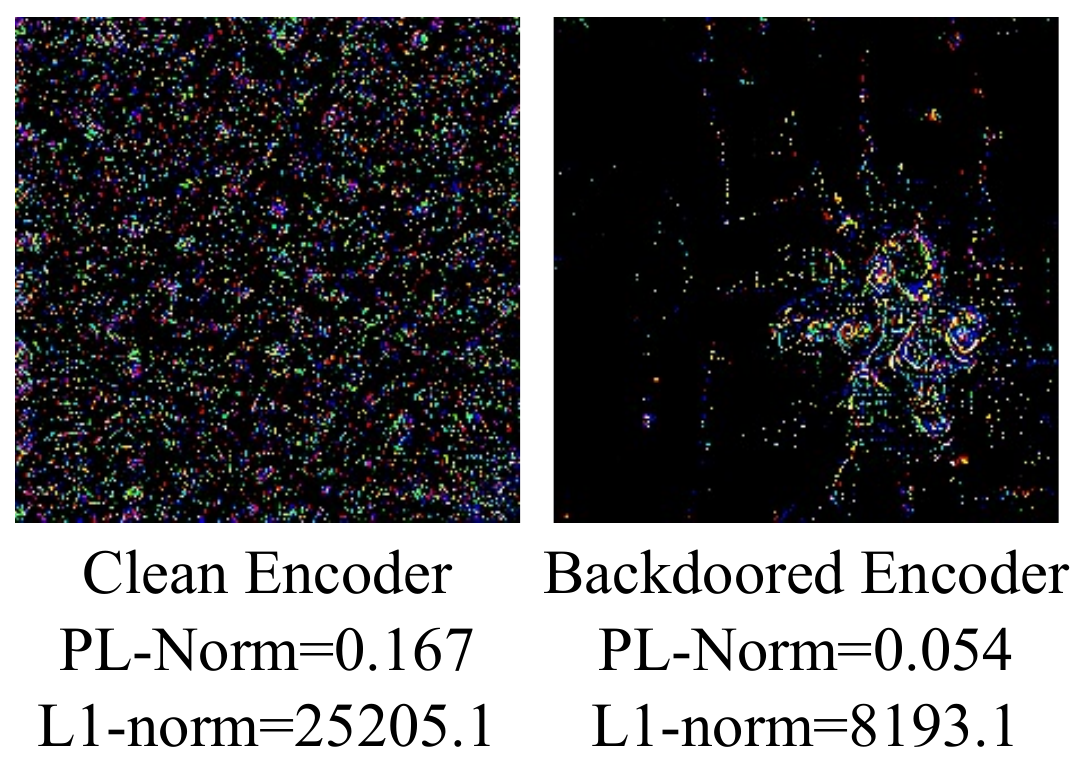}
        \caption{CLIP-Text}
        \label{fig:inverted_imagenet_rn50}
    \end{minipage}
\end{subfigure}

\caption{Inverted Triggers.
    Subfigures~\ref{fig:c10rn18}~\ref{fig:c10rn34}~\ref{fig:c10rn50}~\ref{fig:imgnet-rn50} are {\it Image-on-Image} attacks. Subfigure~\ref{fig:clip-img} is {\it Image-on-Pair} attack. Subfigure~\ref{fig:clip-text} is {\it Text-on-Pair} attack.
    Note that our goal is to do detection and thus it is not that necessary to invert exactly the same trigger as the injected one.
    \toolname is effective at detection since it quantitatively leverages the proposed metric \metricname to decide whether the given encoder is backdoored or not. 
    Visually, triggers inverted from backdoored encoders share common features with ground truth triggers, as they tend to cluster and shift towards the corner while those inverted from clean encoders are evenly distributed throughout the entire image. 
}
\label{fig:trigger}

\end{figure*}

In Figure~\ref{fig:trigger}, we show the triggers inverted by \toolname. The ground truth trigger is a white square located at the right bottom of the image. For Figure~\ref{fig:c10rn18}~\ref{fig:c10rn34}~\ref{fig:c10rn50}, the ground truth trigger shape (height, width, channel) is (10, 10, 3). For Figure~\ref{fig:imgnet-rn50}~\ref{fig:clip-img}~\ref{fig:clip-text}, the ground truth trigger shape (height, width, channel) is (24, 24, 3). 

For each setup, we show a trigger inverted from clean encoder, and a trigger inverted from backdoored encoder. We also report the value of \metricname for each trigger in the figure. 
Notice that (1) triggers inverted from backdoored encoders exploit significantly less pixels than those inverted from clean encoders, and thus their \metricname are lower, (2) triggers inverted from backdoored encoders tend to cluster and shift towards the corner, while those inverted from clean encoders are likely to evenly distribute throughout the entire image. 
For example, in Figure~\ref{fig:inverted_cifar10_rn18}, the trigger from clean encoder scatters over almost the whole image, while the trigger from the backdoored encoder centralizes at the lower right part of the image. 
One can still make similar observations under {\it Text-on-Pair} attack. Take Figure~\ref{fig:inverted_imagenet_rn50} as an example. The trigger from clean encoder evenly distributes across the image, while the trigger from backdoored encoder densely distributes in the lower right region.

\begin{table*}[t!]
  \centering
  \footnotesize
  \vspace{-10pt}
  \caption{Detection time consumed by existing backdoor scanners and our \toolname}
  \label{tab:runtime}
  \begin{tabular}{ccccccccc}
    \toprule
    \multirow{2}{*}{Network} &
    \multicolumn{2}{c}{Training Classifier} &
    \multicolumn{2}{c}{Neural Cleanse} &
    \multicolumn{2}{c}{ABS} &
    \multicolumn{2}{c}{DECREE} \\
    \cmidrule(lr){2-3}\cmidrule(lr){4-5}\cmidrule(lr){6-7}\cmidrule(lr){8-9}
    ~      & ASR & Time (m) & FN & Time (m) & FN & Time (m) & FN & Time (m) \\
    \midrule
    ResNet18   & 1.0   & 64.66 $\pm$ 10.30     & 0 & 4.75 $\pm$ 0.45  & 0 & 2.80 $\pm$ 0.04 & 0 & 0.26 $\pm$ 0.01 \\
    ResNet34   & 1.0   & 63.99 $\pm$ 10.33     & 1 & 9.71 $\pm$ 1.44  & 0 & 5.52 $\pm$ 0.87 & 0 & 0.33 $\pm$ 0.01\\
    \bottomrule
  \end{tabular}
\end{table*}

\section{ROC of \toolname on Different Datasets}
\label{sec:app:roc}

\begin{figure}[!h]
        \centering
        \vspace{5pt}
        \includegraphics[width=.38\textwidth]{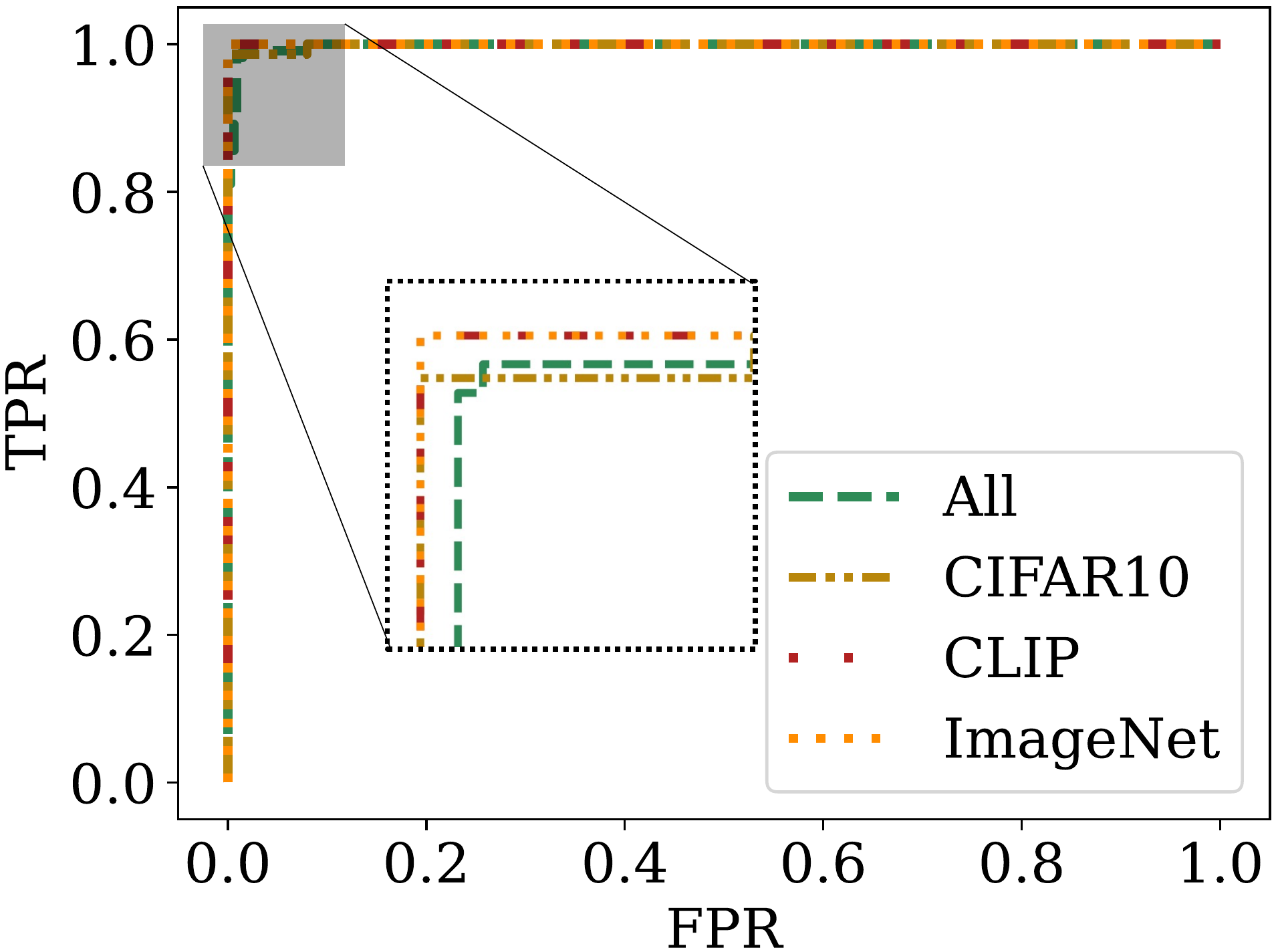}
        \caption{ROC of Detection}
        \label{fig:roc}
\end{figure}

We further use the ROC (Receiver Operating Characteristic) to quantify the effectiveness of our detection method. Given a set of encoders, \toolname inverts triggers from each of them and computes \metricname. After that, to distinguish the backdoored encoders from the benign ones, one can set a threshold for \metricname. The ROC curves are shown in Figure~\ref{fig:roc}. These curves depict how the True Positive Rate (TPR, marked by the vertical axis) and False Positive Rate (FPR, marked by the horizontal axis) change when different thresholds are selected. The green curve denotes the ROC obtained on all the 444 encoders. That is, we set one universal threshold for all the setups, regardless of the architectures of encoders or the dimensions of data samples. We can see that the TPR increases sharply with an almost zero FPR. It achieves an AUC of 0.998, which indicates \metricname effectively distinguishes benign encoders from backdoored encoders without any knowledge about specific setups. Thus \toolname is generally effective on different encoders and different datasets.
Moreover, if we have the knowledge about the pre-training dataset, which is a reasonable assumption in the real-world scenario, the AUC further improves to 0.999 for CIFAR10 and 1.000 for ImageNet and CLIP. Their ROC are depicted by brown, red, and orange curves, respectively.

\section{Time Efficiency}
\label{sec:app:efficiency}

We evaluate the efficiency of \toolname in comparison with two SOTA backdoor scanning techniques, i.e., Neural Cleanse (NC)~\cite{wang2019neural} and ABS~\cite{liu2019abs}.
For both ResNet18 and ResNet34 architectures, we conduct experiments on 10 backdoored encoders pre-trained on CIFAR10. The attack target is a “one” image from the attack dataset SVHN.

Note that \toolname is an order of magnitude faster than the other two baselines, even without considering the training time for downstream classifiers.
This is because \toolname generates just one trigger for each encoder and do not have to scan each label like what NC and ABS do. In addition, we find that NC have one False Negative during the experiment, further validating the necessity and motivation of our \toolname.

\section{Adaptive Attack} \label{sec:app:adaptive_attack}
In addition to existing attacks, We design an adaptive attack, as explained in Section~\ref{sec:eval:adaptive_attack}. $\alpha$ in Eq.~\ref{eq:adapt_attack_loss} is a hyper-parameter that controls the cosine similarity loss during the attack. Intuitively, when $\alpha$ becomes larger, the images stamped with trigger will share less similar embeddings. When $\alpha$ is near to zero, the images with trigger tend to have extremely similar embeddings, which also means they are similar to the embedding of the attack target. For different $\alpha$ values, we train 10 trojaned encoders and show their average metrics in Table~\ref{tab:adaptive_attack}. The encoders are pre-trained on CIFAR10 with ResNet18 architecture and the attack target is a ``truck'' image from the attack dataset STL-10.
\begin{table}[t!]
  \centering
  \footnotesize
  \caption{Encoders Adaptively Attacked by Eq.~\ref{eq:adapt_attack_loss}}
  \label{tab:adaptive_attack}
  \begin{tabular}{lcccc}
    \toprule
    ~ & Accuracy & ASR & $\normlone$-Norm & \metricname \\
    \midrule
    $\alpha=0$ & 76.22 & 99.73 & 171.65 & 0.056 \\
    $\alpha=0.5$ & 72.95 & 93.60 & 258.57 & 0.084 \\
    $\alpha=1.0$ & 72.48 & 69.90 & 430.08 & 0.140 \\
    $\alpha=2.0$ & 72.08 & 31.00 & 847.45 & 0.276 \\
    \bottomrule
  \end{tabular}
\end{table}

According to Table~\ref{tab:adaptive_attack}, \toolname stays effective when $\alpha=0.5$, as encoders with \metricname $<0.1$ are detected as trojaned. When $\alpha$ further increases, the adaptive attack evades our detection. However, the ASR drops a lot at the same time, from over 90\% to below 70\%, even around 30\%. Therefore, it is quite difficult for the attackers to evade our detection with a high ASR.

\section{Ablation Study}
\label{app:abl}

This section studies the effectiveness of \toolname against different trigger patterns and sizes. We also studies the impact of hyper-parameters. The results show that \toolname has a robust design.

\subsection{Different Trigger Patterns and Sizes} \label{sec:app:different_triggers}

\noindent
\textbf{Trigger Configurations.} We test the effectiveness of \toolname on triggers with different configurations. The experimental results are shown in Table~\ref{tab:trigger_config}. Encoders with \metricname < 0.1 are detected as trojaned.
The default trigger pattern is a 10$\times$10 white square located at lower-right corner. \looseness=-1

We can see that \toolname effectively inverts relatively small triggers for all encoders trojaned by triggers with different colors, positions, and textures. That means \toolname can successfully detect trojaned encoders in different trigger patterns. We also show the effectiveness of \toolname against different trigger size in Table~\ref{tab:trigger_size}. 

\begin{table}[h!]
    \centering
    \footnotesize
    \caption{Detection Results on Different Trigger Patterns. We alter the configurations of triggers and conduct {\it Image-on-Image} attacks with them. The 1-2 columns are the configurations we change. The 3-4 column are the \lonenorm  \ and \metricname of inverted triggers generated by \toolname. For each row, we evaluate on 5 encoders and compute the average. All the encoders are pre-trained on CIFAR10 and the attack target is an image of label {\it one} from SVHN.
    }
    \label{tab:trigger_config}
    \begin{tabular}{rlrr}
         \toprule
         Config. & Value & \lonenorm & \metricname\\
         \midrule
         \multirow{3}{*}{Color} &   Green   & 250.43    & 0.082\\
                                &   Purple  & 248.48    & 0.081\\
                                &   White   & 113.99    & 0.037\\
        \midrule
        \multirow{3}{*}{Position}   &   Lower-Right &   113.99  &   0.037   \\
                                    &   Center      &   135.84  &   0.044   \\
                                    &   Upper-Left  &   123.72  &   0.040   \\
        \midrule
        \multirow{3}{*}{Texture}    &   Random  &   50.09   &   0.016\\
         &   TrojanNN~\cite{liu2017trojaning} &  58.30    &   0.019\\
         &   White   &   113.99  &   0.037\\
        \bottomrule
    \end{tabular}
\end{table}

\begin{table}[h!]
    \centering
    \footnotesize
    \caption{Detection Results on Different Trigger Sizes. The input image size of encoders is 32$\times$32. 
    }
    \label{tab:trigger_size}
    \begin{tabular}{lrr}
        \toprule
        Trigger Size (Ratio) & \lonenorm & \metricname\\
         \midrule
        5$\times$5 (2.4\%)   &   36.44   &   0.012\\
        7$\times$7 (4.8\%)     &   44.38   &   0.014\\
        10$\times$10 (9.8\%)  &   113.99  &   0.037\\
        12$\times$12 (14.0\%)  &   135.19  &   0.044\\
        14$\times$14 (19.1\%)  &   150.76    &   0.049\\
        \bottomrule
    \end{tabular}
\end{table}

\subsection{Hyper-parameters}\label{sec:app:hyper_params}

\noindent {\bf Effect of shadow dataset size $M$.}
In our evaluation, we use shadow dataset (containing 1000 images) to do trigger inversion. We further evaluate on smaller shadow dataset to show that \toolname is not sensitive to the shadow dataset size $M$, as shown in the Table~\ref{tab:impact_shadow_dataset}. 
Note that encoders with \metricname < 0.1 are detected as trojaned.

\begin{table}[h!]
  \centering
  \footnotesize
  \caption{Impact of Shadow Dataset Size $M$. Encoders are trained on CIFAR10 and shadow dataset are randomly sampled from CIFAR10. We keep batch size $N$ to be 128 during self-supervised trigger inversion.}
  \label{tab:impact_shadow_dataset}
  \begin{tabular}{lrrr}
    \toprule
    $M$ & 50 & 100 & 1000 \\
    \midrule
    \lonenorm & 105.2 & 106.59 & 113.99 \\
    \metricname & 0.034 & 0.035 & 0.037 \\
    \bottomrule
  \end{tabular}
\end{table}

\noindent {\bf Effectiveness of threshold $\tau$.}
We assign a pre-defined value to $\tau$ = 0.1. 
We further clarify that $\tau$ = 0.1 is sufficient to do effective detection.

As shown in the Table~\ref{tab:trigger_size}, we evaluate on 5 different sizes of triggers, the ratio of which ranging from 2.5\% to 20\%. All of these triggers have a \metricname < 0.1 because the encoder just learns part of the trigger feature during the trojaning procedure. 
Additionally, any trigger with a larger ratio than 20\% (occupying almost a quarter of the whole image) is not a reasonable trigger since this violate the principle of stealthiness for attackers. 
Therefore, $\tau$ = 0.1 is a reasonable upper-bound for trigger size ratios and thus an effective threshold for \toolname.

\section{Advanced Attacks} \label{sec:app:advanced_attack}

Existing backdoor attacks on self-supervised learning are only effectively conducted when using patch-based sample-agnostic triggers~\cite{jia2022badencoder}~\cite{carlini2022poisoning}. 

To provide better understanding of backdoor attack against self-supervised learning,
we adapt 2 existing ``advanced attacks'' (image-size and sample-specific attacks) from supervised learning into our settings, namely WaNet~\cite{nguyen2020wanet} and Invisible~\cite{Li2021invisible}. 
We follow the attack procedure of BadEncoder~\cite{jia2022badencoder}, the {\it Image-on-Image} attack we have adopted in our paper, and only change the trigger pattern from patch-based triggers to image-size triggers generated by WaNet and Invisible. Then we evaluate ASR on the downstream classifier trained from the trojaned encoder. The results is shown in Table~\ref{tab:advanced_attack}. \looseness=-1

\begin{table}[h!]
  \centering
  \footnotesize
  \caption{\revise{Advanced Attacks}. ASR is evaluated on the downstream classifiers trained on STL-10. The encoders are pre-trained on CIFAR10 with ResNet18 architecture and the attack target is a “truck” image from the attack dataset STL-10.}
  \label{tab:advanced_attack}
  \begin{tabular}{lccc}
    \toprule
      & WaNet & Invisible & BadEncoder \\
    \midrule
     ASR & 10.23 & 10.02 & 99.73\\
    \bottomrule
  \end{tabular}
\end{table}

From the experimental result, we can observe that image-size and sample-specific backdoor attacks can hardly be successful on self-supervised learning pre-trained encoders. 
These attacks can be successful and stealthy in supervised learning because there exist a concrete target label that can enable a strong hint during attacking. 
However, self-supervised learning only consider positive or negative pairs. Without distinct and obvious features (like patch-based triggers), such sample-specific triggers can hardly establish a strong correlation between victim images and target images. 

\section{More SSL Attacks} \label{sec:app:ssl_attack}

We study on 3 emerging SSL attacks, namely SSLBackdoor~\cite{Saha2022backdoorSSL},
CorruptEncoder~\cite{zhang2023corruptencoder} and CTRL~\cite{li2022demystifying}.

Our method successfully detected CorruptEncoder with \metricname of approximately 0.08 but failed to identify SSLBackdoor and CTRL, both of which had \metricname around 0.23.
The reason for our failure to detect SSLBackdoor was its low ASR (<10\%), which falls outside of our expected ASR range (>99\%), as stated in our threat model. 
Although SSLBackdoor had good false positive scores, its stealthy nature made it difficult to detect. 
Our method also failed to detect CTRL since it used a pervasive trigger that was outside of our threat model (patch-like triggers).

\end{document}